\pdfoutput=1
\documentclass[11pt]{article}

\usepackage[]{acl}

\usepackage{times}
\usepackage{latexsym}

\usepackage[T1]{fontenc}
\usepackage[utf8]{inputenc}

\usepackage{microtype}

\usepackage{graphicx}
\usepackage{booktabs}
\usepackage{caption}
\usepackage{subcaption}
\usepackage{marvosym}
\usepackage{mathtools}
\usepackage{amsmath}
\usepackage{amssymb}
\usepackage{comment}
\usepackage{adjustbox}
\usepackage{lipsum}
\usepackage{xspace}
\usepackage{longtable}
\usepackage{multirow}
\usepackage{multicol}
\usepackage{makecell}
\usepackage{enumitem}
\usepackage{arydshln}

\usepackage{tikz}
\usepackage{tikz-qtree}
\usepackage{float}

\usetikzlibrary{arrows}
\usetikzlibrary{calc}
\usetikzlibrary{decorations.pathreplacing}
\usetikzlibrary{positioning}
\title{Prefix-to-SQL: Text-to-SQL Generation from Incomplete User Questions}

\author{
{\Large Naihao Deng$^{1}$
\quad Shuaichen Chang$^{2}$
\quad Peng Shi$^{3}$
\quad Tao Yu$^{4}$
\quad Rui Zhang$^{5}$
}\\
{\large
$^1$ The University of Michigan, Ann Arbor \quad $^2$ The Ohio State University}\\
{\large
\quad $^3$ The University of Waterloo
\quad $^4$ The University of Hong Kong  \quad $^5$ Penn State University}\\
{\small \tt dnaihao@umich.edu,
 chang.1692@osu.edu,
 peng.shi@uwaterloo.ca,
 tao.yu@yale.edu,
 rmz5227@psu.edu}
}

\newcommand{\datas}{{PAGSAS}\xspace}
\newcommand{\task}{{prefix-to-SQL}\xspace}

\def\SB#1{\textsubscript{{#1}}}
\def\SPSB#1#2{\rlap{\textsuperscript{{#1}}}\SB{#2}}

\begin{document}

\maketitle

%%%%%%%%%%%%%%%%%%%%%%%%%%%%%%%%%%%%%%%%%%%%%%%%%%%%%%%%%%%%%%%%%%%%%%%%%%%%%%%%%%%%%%%%%%%%%%%%%%%%%%%%%%
%%%%%%%%%%%%%%%%%%%%%%%%%%%%%%%%%%%%%%%%%%%%%%%%%%%%%%%%%%%%%%%%%%%%%%%%%%%%%%%%%%%%%%%%%%%%%%%%%%%%%%%%%%

%  abstract

\begin{abstract}
Existing text-to-SQL research only considers complete questions as the input, but lay-users might strive to formulate a complete question. To build a smarter natural language interface to database systems (NLIDB) that also processes incomplete questions, we propose a new task, \textbf{\task} which takes question prefix from users as the input and predicts the intended SQL. We construct a new benchmark called \textbf{\datas} that contains 124K user question \textbf{p}refixes and the intended SQL for 5 sub-tasks \textbf{A}dvising, \textbf{G}eoQuery, \textbf{S}cholar, \textbf{A}TIS, and \textbf{S}pider. Additionally, we propose a new metric \textsc{Save} to measure how much effort can be saved by users. Experimental results show that \datas is challenging even for strong baseline models such as T5. As we observe the difficulty of \task is related to the number of omitted tokens, we incorporate curriculum learning of feeding examples with an increasing number of omitted tokens. This improves scores on various sub-tasks by as much as $9\%$ \textsc{Recall} and $3.9\%$ \textsc{Save} scores on sub-task GeoQuery in \datas.
\end{abstract}
%%%%%%%%%%%%%%%%%%%%%%%%%%%%%%%%%%%%%%%%%%%%%%%%%%%%%%%%%%%%%%%%%%%%%%%%%%%%%%%%%%%%%%%%%%%%%%%%%%%%%%%%%%
%%%%%%%%%%%%%%%%%%%%%%%%%%%%%%%%%%%%%%%%%%%%%%%%%%%%%%%%%%%%%%%%%%%%%%%%%%%%%%%%%%%%%%%%%%%%%%%%%%%%%%%%%%

% Introduction

\section{Introduction}
\label{sec: introduction}

Text-to-SQL aims to translate natural utterances to executable SQL queries in relational databases. Effective natural language interfaces to databases (NLIDB) give lay-people access to vast amounts of data stored in relational databases~\cite{finegan2018improving}. However, existing text-to-SQL research only considers complete questions as the input. Users might struggle to formulate proper questions to retrieve their desired information~\cite{sordoni2015hierarchical,mccamish2018data}.

% AutoSQL framework
\begin{figure}[ht!]
    \centering
    \includegraphics[width=0.48\textwidth]{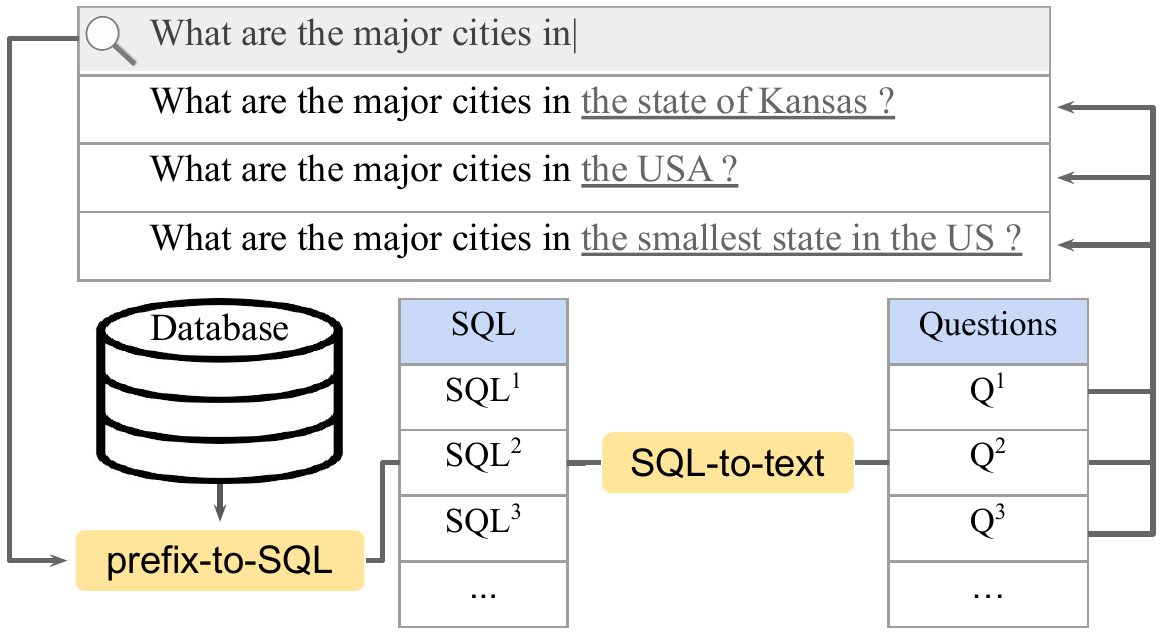}
    \caption{A more user-friendly NLIDB. Given users' incomplete input, this NLIDB can generate possible SQL queries and their corresponding questions.}
    \label{fig:framework}
\end{figure}

% Examples of missing entities in dataset
\begin{table*}[ht!]
        \includegraphics[width=\textwidth]{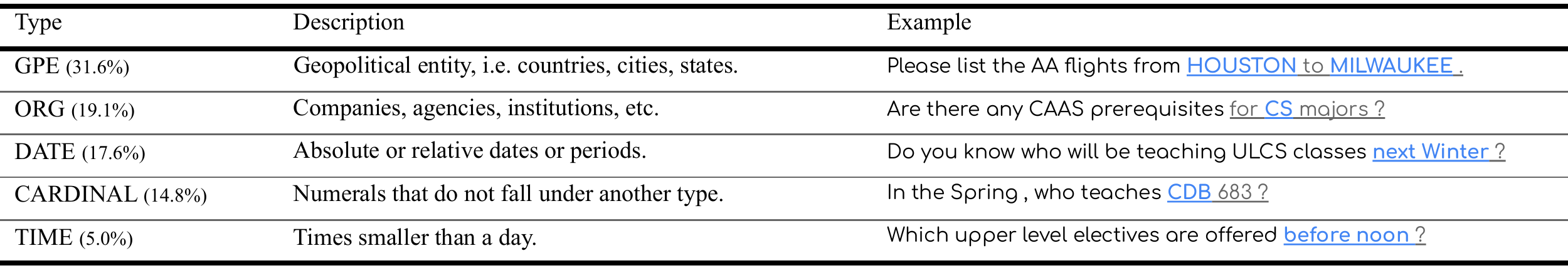}
    \caption{Examples of different types of omitted entities in \datas. In the ``Example'' column, the \underline{underlined} text is omitted to generate the question prefix. The blue text represents the corresponding entities. In the ``Type'' column, the number in the bracket represents the percentages of omitted entity types in \datas. Here we only report entity types with a frequency $\ge5.0\%$.}
    \label{tab: missing-ent-example}
    \vspace{-3mm}
\end{table*}

To build a more user-friendly NLIDB model as shown in Figure~\ref{fig:framework}, we propose a new task, \textbf{\task} which takes the prefix of user questions and predicts the intended SQL for the database system. For the second stage of SQL-to-text, we collect canonical rules for converting the intended SQL queries to question texts. Users can quickly choose one of the suggested question completions or alternatives and thus do not have to type the whole question by themselves~\cite{bhatia2011query, shokouhi2013learning, cai2016query}. The rules guarantee that the suggested question matches the SQL query~\cite{yao2019model}. Thus the executed SQL query is what the user intends. 

We focus on \task in Figure~\ref{fig:framework} in this work. To simplify and better define \task, we make three assumptions: 
(1) Examples from the original text-to-SQL dataset represent how database users are ``likely'' to query the database. We consider these SQL queries as ``correct'' SQL predictions for \task because we think users would have similar intentions to the existing database users.
(2) We consider SQL queries that are not in the original text-to-SQL dataset as ``incorrect'' because they are ``unlikely'' SQL queries when users query the database. In other words, we do not want our models to cover all possible combinations of tables, columns, and values for SQL. Instead, models need to learn how users are ``likely'' to query the database in their daily usage.
(3) Following query auto-completion works in information retrieval~\cite{sordoni2015hierarchical,bhatia2011query}, we consider the question prefix to be the primary incomplete question type and work on \task in this work.

For \task, we build the \textbf{\datas} benchmark, which consists of 5 sub-tasks and 124K examples from Advising~\cite{finegan2018improving}, GeoQuery~\cite{10.5555/1864519.1864543}, Scholar~\cite{iyer-etal-2017-learning}, ATIS~\cite{price1990evaluation,dahl1994expanding} and Spider~\cite{yu-etal-2018-spider} (Examples in Table~\ref{tab: missing-ent-example}). To evaluate the performance of our baseline models on \datas, we present \textsc{Recall} and \textsc{MRR} scores for how many SQL predictions are correct and how high the model ranks the first correct SQL query, respectively. Additionally, we propose a new metric, \textbf{\textsc{Save}} which measures how much user effort can be saved by models from the user's perspective. 

Our results show that both generative and retrieval models can predict SQL queries that match the user's intention, but the scores are much lower than the original text-to-SQL task~\cite{finegan2018improving, dong2018coarsetofine, yin2018tranx, cao-etal-2019-semantic, rubin2021smbop, huang2021relation, cao2021lgesql, shi2020learning, zhao2021gp, yu2021grappa}. \datas is challenging even for strong baseline models such as T5. By analysis of models' performance, we find \task differs from the original text-to-SQL task. For the original text-to-SQL task, short SQL queries are considered easier than longer ones \cite{finegan2018improving}. In contrast, models' performance is negatively correlated with the number of omitted tokens for \task. Thus,  \task poses new challenges compared to the existing text-to-SQL.

Based on our analysis, we adopt curriculum learning by feeding examples with an increasing number of omitted tokens. This significantly improves the model's performance on various sub-tasks in \datas. On sub-task GeoQuery, T5 with curriculum learning improves \textsc{Recall} and \textsc{Save} by as much as $9\%$ and $3.9\%$, respectively.

% Summarized the contribution of this paper 
In summary, our contributions are three-fold:
\begin{itemize}
    \item We introduce a new task \task with the \datas dataset that predicts SQL queries based on their question prefix. We design a new metric \textsc{Save} and evaluate baselines' performance with metrics \textsc{Recall}, \textsc{MRR} and \textsc{Save} on \datas.
    \item Experiments show that \datas is challenging even for strong text-to-SQL baseline models such as T5. We analyze the baseline results and demonstrate that \task poses new challenges to existing models.
    \item We propose curriculum learning based on number of omitted tokens in thep prefix, which improves scores on various sub-tasks by as much as $9\%$ \textsc{Recall} scores on sub-task GeoQuery in \datas.
\end{itemize}
 
 \begin{table*}[ht!]
 
 \small
 \setlength{\tabcolsep}{8pt}
 \centering
     \begin{tabular}{l r r r r r r r r r r r r r r}
     \toprule
      & \multicolumn{2}{c}{Advising} & \multicolumn{2}{c}{ATIS} & \multicolumn{2}{c}{GeoQuery} & \multicolumn{2}{c}{Scholar} & \multicolumn{2}{c}{Spider}  \\
       
      & \multicolumn{1}{c}{Q} & \multicolumn{1}{c}{S} & \multicolumn{1}{c}{Q} &  \multicolumn{1}{c}{S} & \multicolumn{1}{c}{Q} & \multicolumn{1}{c}{S} & \multicolumn{1}{c}{Q} & \multicolumn{1}{c}{S} & \multicolumn{1}{c}{Q} & \multicolumn{1}{c}{S}  \\
     \midrule
      \# prefixes (P) ($10^3$)  & 30.7 & 38.8 & 34.2 & 33.4 & 3.1 & 3.0  & 4.6 & 4.4 & 51.4 & 51.1 \\
      \# SQL queries (S) ($10^3$) & 39.7 & 51.2  & 56.9 & 56.4  & 6.8 & 6.8  & 6.2 & 6.0 & 63.0 & 62.9 \\
      S/P & 1.3 & 1.3 & 1.7 & 1.7 & 2.2 & 2.2 & 1.3 & 1.4 & 1.2 & 1.2\\
      $\mu$ tokens in prefix & 7.6 & 7.7 & 8.9 & 9.0 &6.6 & 6.6 & 5.6 &  5.6 & 9.7 & 9.7\\
      $\mu$ tokens in complete questions & 12.1 & 12.1 & 12.6 & 12.6 & 8.3 & 8.3 &8.5 & 3.8 &15.9 & 15.9\\
      
      $\mu$ omitted tokens & 5.6 & 5.7 & 5.9 & 6.0 & 3.7 & 3.7 & 3.8 & 3.8 & 7.5 & 7.5\\
      
     \bottomrule
     
     \end{tabular}
 \caption{Statistics for each domain in \datas on question splits (``Q'' columns) and SQL query splits (``S'' columns).
 ``\#'' means ``the number of'', ``$\mu$'' means ``the average number of''. Because each SQL query corresponds to a complete question, \# complete questions = \# SQL. Note that $\mu$ omitted tokens $\neq$ $\mu$ tokens in complete questions - $\mu$ tokens in prefix, because we group prefix as described in Section~\ref{subsec: dataset-construction}.}
 \label{tab:data-stat-all}
 \end{table*}
 
%%%%%%%%%%%%%%%%%%%%%%%%%%%%%%%%%%%%%%%%%%%%%%%%%%%%%%%%%%%%%%%%%%%%%%%%%%%%%%%%%%%%%%%%%%%%%%%%%%%%%%%%%%
%%%%%%%%%%%%%%%%%%%%%%%%%%%%%%%%%%%%%%%%%%%%%%%%%%%%%%%%%%%%%%%%%%%%%%%%%%%%%%%%%%%%%%%%%%%%%%%%%%%%%%%%%%

% Related works

\section{Related Work}

\paragraph{Question Auto-completion}
Our task is inspired by the task of question auto-completion (QAC), but target at predicting SQL queries based on the incomplete question. QAC refers to that when the user gives a prefix, the user interface proposes alternative ways of extending the prefix to a full question query~\cite{cai2016query}. This task is also known as type-ahead~\cite{xiao2013efficient,cai2014time,li2009efficient,li2011efficient} or auto-complete suggestion~\cite{jain2010organizing}.
Early work on QAC primarily focused on word prediction~\cite{vanderheiden1987comparative,swiffin1987adaptive,darragh1991adaptive}, while sentence completion received more attention later~\cite{10.1145/1008992.1009066,bickel2005learning,nandi2007effective}. The task of sentence completion refers to that when the user gives a sentence's initial fragment, the system identifies the remaining part of the sentence that the user intends to write~\cite{cai2016query}.
Heuristic approaches~\cite{bar2011context,cai2014time,10.1145/2348283.2348364,zhang2015adaqac,whiting2014recent} as well as learning-based approaches~\cite{cai2016learning,jiang2014learning,mitra2015exploring} has been proposed for QAC.
\paragraph{Natural Language Interfaces to Databases}
The task of building natural language interfaces to database (NLIDB) has aroused great interest in both the NLP and DB communities~\cite{warren1981efficient,androutsopoulos1995natural,popescu2004modern,hallett2006generic,giordani2012generating}. In the NLP community, mapping natural language utterances to SQL queries for databases, also known as SQL-based semantic parsing, has attracted increasing attention~\cite{yin2017syntactic,dong2018coarse,guo2019towards,rat-sql,bogin2019representing,lin2020joint,yin2019reranking}.
Early work of NLIDB primarily focuses on~\cite{hemphill1990atis,dahl1994expanding,zelle1996learning}. Benchmarks such as Academic~\cite{li2014constructing}, IMDB and Yelp~\cite{yaghmazadeh2017sqlizer} are too small and contain fewer than 200 questions. Restaurants~\cite{tang-mooney-2000-automated} contains 23 unique SQL queries, much fewer than other benchmarks such as GeoQuery (246 unique SQL queries)~\cite{finegan2018improving}. Thus, we do not include Academic, IMDB, Yelp, or Restaurants in our constructed dataset. Recently, cross-domain datasets including Spider~\cite{yu-etal-2018-spider} and WikiSQL~\cite{zhongSeq2SQL2017} are proposed to evaluate model's ability of generalization to unseen domains~\cite{suhr2020exploring}. However, WikiSQL contains only simple SQL queries and single tables, which is too simple~\cite{yu-etal-2018-spider}. Thus, we do not construct our benchmark from WikiSQL.

%%%%%%%%%%%%%%%%%%%%%%%%%%%%%%%%%%%%%%%%%%%%%%%%%%%%%%%%%%%%%%%%%%%%%%%%%%%%%%%%%%%%%%%%%%%%%%%%%%%%%%%%%%
%%%%%%%%%%%%%%%%%%%%%%%%%%%%%%%%%%%%%%%%%%%%%%%%%%%%%%%%%%%%%%%%%%%%%%%%%%%%%%%%%%%%%%%%%%%%%%%%%%%%%%%%%%

% Task Definition and Dataset Construction

\section{Task Definition and Dataset Construction}

We will focus on \task as presented in Figure~\ref{fig:framework}. For the second stage SQL-to-text, we present some of the rules in Table~\ref{tab:canonical-question} in Appendix~\ref{sec:appendix-canonical}.

\subsection{Dataset Construction}
\label{subsec: dataset-construction}

\begin{figure*}[ht!]
    \centering
   \begin{subfigure}[c]{0.3\textwidth}
        \includegraphics[width=\textwidth]{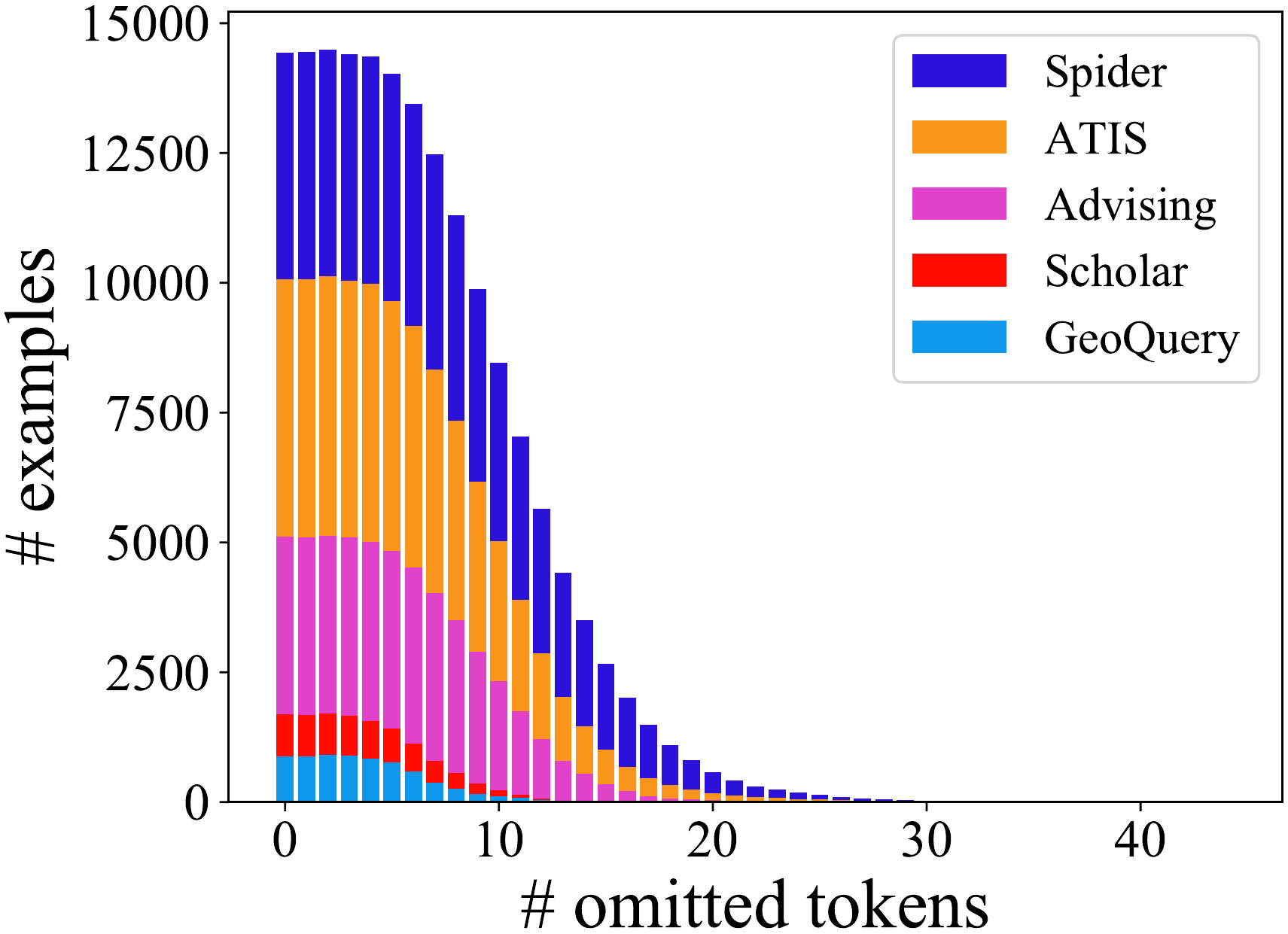}\caption{}
        \label{subfig:question_missing_len_distribution}
    \end{subfigure}\hspace{5mm}
    ~ 
    \begin{subfigure}[c]{0.3\textwidth}
         \includegraphics[width=\textwidth]{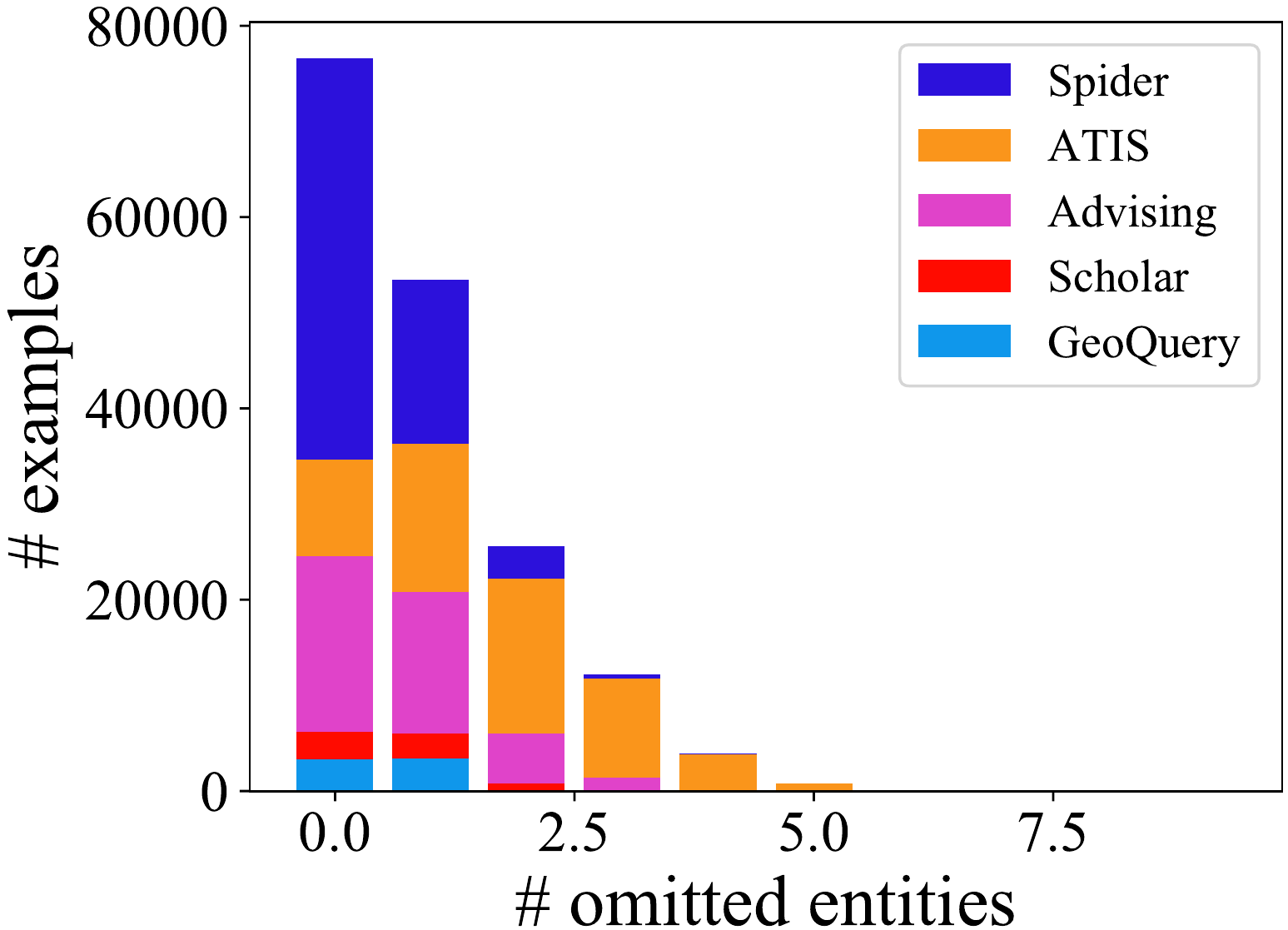}
         \caption{}
        \label{subfig:question_missing_ents}
    \end{subfigure}\hspace{5mm}
    ~
    \begin{subfigure}[c]{0.3\textwidth}
         \includegraphics[width=\textwidth]{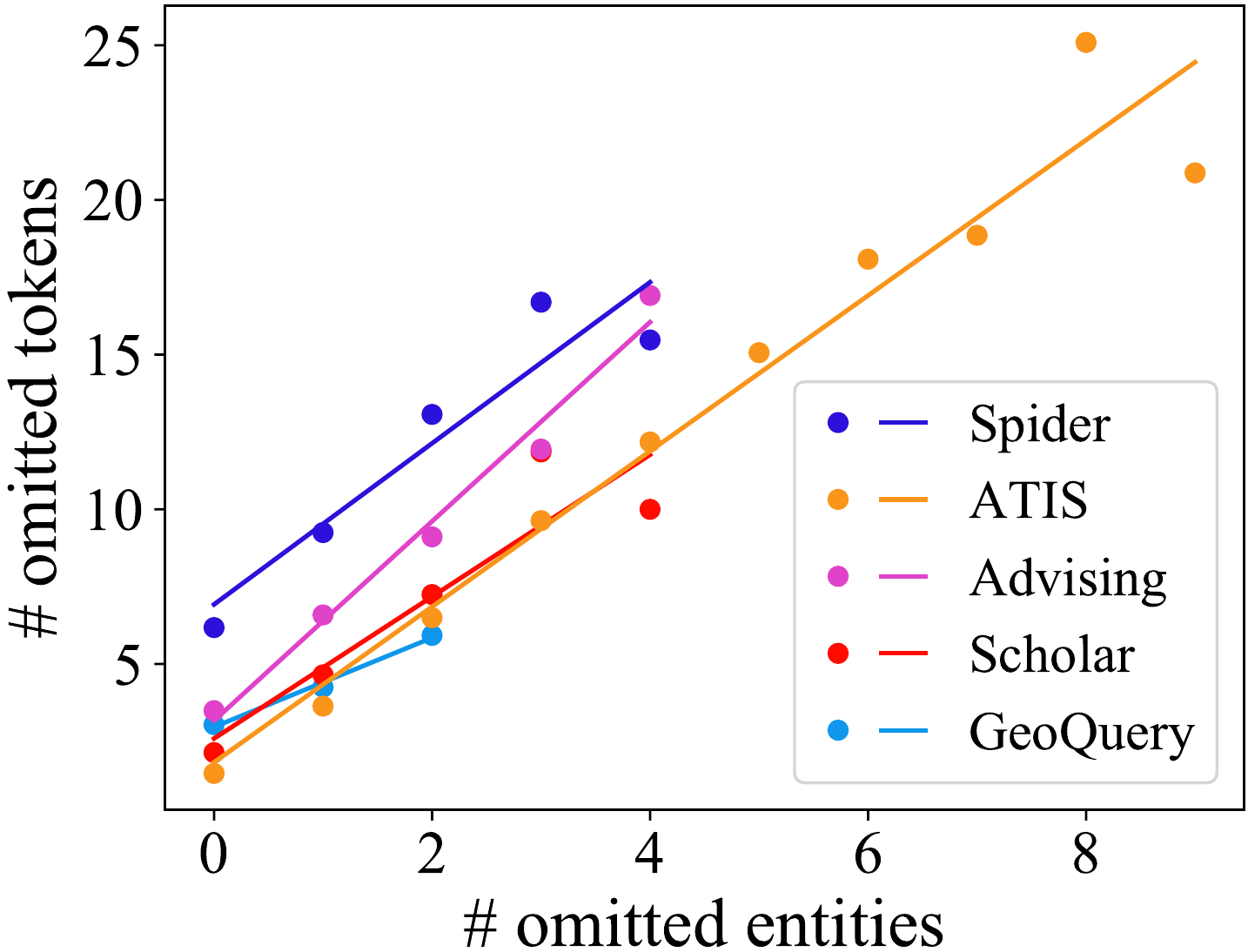}
         \caption{}
        \label{subfig:question_omit_len_ent_ratio}
    \end{subfigure}\hspace{5mm}
    
    \caption{Stacked distribution for (a) the number of omitted tokens from complete questions and (b) the number of omitted entities for question split; (c) the number of omitted tokens (vertical axis) v.s. the number of omitted entities (horizontal axis) for question split. The horizontal axis represents (a) the number of omitted tokens, (b) the number of entities in the omitted text for each sub-task in \datas. The vertical line in (a) and (b) represents the corresponding number of examples.}
    \label{fig: question-missing-len-info}
\end{figure*}

\paragraph{Construction Methodology}

Following query auto-completion in information retrieval~\cite{park2017neural,bar2011context,krishnan2020generation}, we generate question prefixes from the first to the last token of each question from the existing text-to-SQL datasets. Note that prefixes spanning to the last token represent the complete question. We include this because users might go ahead and type their entire question in actual usages of our system.

For a question prefix $q^{\text{pre}}$, we assume the complete question $q^{\text{cmp}}$ represents the user's intention, and the corresponding gold SQL $s^{\text{gold}}$ is the SQL representation of the user's intention.
Note that a question prefix can extend to multiple complete questions, so we group gold SQL queries if their corresponding complete questions share the same question prefix. We treat these SQL queries as the gold SQL predictions for the prefix following our first and second assumptions in Section~\ref{sec: introduction}. Therefore, each example in our dataset contains a question prefix $q^{\text{pre}}$ and its corresponding list of $L$ gold SQL queries $S^{\text{gold}}$ that matches the user's intention:
\begin{equation}
    S^{\text{gold}}=\{s^{\text{gold}}_1, s^{\text{gold}}_2, \dots, s^{\text{gold}}_L\}
\end{equation}

\paragraph{Source Text-to-SQL Datasets}
We construct \textbf{\datas} by \textbf{P}refixes for \textbf{A}dvising, \textbf{G}eoQuery,  \textbf{S}cholar, \textbf{A}TIS, and \textbf{S}pider because these large-scale datasets contain user questions collected in different domains and their corresponding SQL queries are manually annotated. As these datasets contain different domain knowledge~\cite{suhr2020exploring}, we treat them as different sub-tasks and perform experiments on each of them separately. We exclude data that is annotated as ``exclude'' by~\citet{finegan2018improving}.

For the original datasets other than Spider, we adopt the question split and the SQL query split following~\citet{finegan2018improving}. Question split splits data based on their complete questions, and SQL query split splits data based on their corresponding SQL templates. Note that question split is a common real-world setting as users tend to concern information about certain fields. For instance, in the original GeoQuery dataset, $66.25\%$ of SQL queries include the column ``\texttt{state\_name}'' from table ``\texttt{state}'' while only $0.11\%$ SQL queries (1 SQL query) include the column ``\texttt{country\_name}'' under the table ``\texttt{river}''. But still, people might come up with questions or type in question prefixes that correspond to an unseen SQL. To evaluate the model under this challenge, we also include SQL query split in our experimental settings.

The original Spider dataset is a large cross-domain text-to-SQL dataset that consists of 10,181 questions and 5,693 unique complex SQL queries across 138 domains \cite{yu-etal-2018-spider}. The primary evaluation for the original Spider dataset is in the cross-database setting, where models are evaluated on examples for databases not seen during training. One of the primary challenges in this setting is the generalization to new database schemas, while in our task, it does not make sense to suggest questions or predict the corresponding SQL query for a completely unseen domain. Following~\citet{shaw2021compositional}, we adopt a setting similar to an alternative setting called the example split in the original dataset \cite{yu-etal-2018-spider} where the databases are shared between train and test examples. Similar to \citet{shaw2021compositional}, we identify examples from training set databases that contain more than 50 examples to ensure sufficient coverage over table and column names in the training data. We then generate two new training, validation, and test splits consisting of 2789 training, 493 validation, and 1094 test examples across 51 databases: a random split (question split) and a split based on SQL template (SQL query split).

Then we use the aforementioned construction methodology to construct \datas for both question split and SQL query split.

\paragraph{Dataset Analysis}
\label{subsec: dataset-analysis}
Table~\ref{tab:data-stat-all} reports the statistics for~\datas. Figure~\ref{subfig:question_missing_len_distribution}, ~\ref{subfig:question_missing_ents} show the stacked distribution of the number of omitted tokens from complete questions and the number of omitted entities detected by SpaCy \cite{spacy2}, respectively. Figure~\ref{subfig:question_omit_len_ent_ratio} demonstrates that there is a linear relationship between the number of omitted tokens and the number of omitted entities, and more entities will be missing if there are more omitted tokens. SQL query split possesses a similar distribution as Figure~\ref{fig: question-missing-len-info} (Figure~\ref{fig: query-missing-len-info} in Appendix~\ref{appendix: dataset-split-analysis}). We also report the distribution of the number of tokens in prefixes as well as the number of tokens in the complete question in Figure~\ref{fig: len-distribution} in Appendix~\ref{appendix: dataset-split-analysis}. 

Table~\ref{tab: missing-ent-example} gives explanations of the omitted entity types detected by SpaCy~\cite{spacy2} and examples in \datas that correspond to entity types with a frequency greater than or equal to $5.0\%$. 

\subsection{Prefix-to-SQL Task Definition and Evaluation}
\label{sec:taskdef}

Given the database schema $c$ and a question prefix $q^{\text{pre}}$, the model generates a list of top $K$ SQL queries that match possible complete questions:
\begin{equation}
    S^{\text{sug}}_{K}=\{s^{\text{sug}}_1,s^{\text{sug}}_2,\dots,s^{\text{sug}}_K\}
\end{equation}
Based on our first and second assumptions in the Section~\ref{sec: introduction}, we will only consider complete questions in the original text-to-SQL dataset as the gold SQL queries.

\paragraph{Evaluation Metrics}

We use \textsc{Recall} to evaluate how many predicted SQL queries match the gold SQL queries. The equation is given as follows:
$$
\begin{aligned}
    & \textsc{Recall}@K= \frac{|S^{\text{gold}}\cap S^{\text{sug}}_{K}|}{|S^{\text{gold}}|}\\
\end{aligned}
$$
Additionally, we report mean reciprocal rank (\textsc{MRR})~\cite{schutze2008introduction} scores to calculate how high the model will rank the first correctly predicted SQL query.

But neither \textsc{Recall} nor \textsc{MRR} can measure how much effort users can save using the system. Thus, we propose \textsc{Save} to measure how well the model can save user effort. For each complete question $q^{\text{cmp}}$, \textsc{Save} calculates at which token (token $t$) the model can start predicting the correct SQL query corresponding to $q^{\text{cmp}}$. Tokens after $t$ in $q^{\text{cmp}}$ are tokens the user does not need to type. When the model does not predict anything correct, \textsc{Save} becomes 0. Formally, we define \textsc{Save} as:

\begin{small}
\begin{equation}
\label{eq: save-formula-positive-S}
\textsc{Save}@K= 
\left.\frac{\text{len}(q^{\text{cmp}}) - \min\left\{\text{len}(q_j^{\text{pre}})\right\}}{\text{len}(q^{\text{cmp}})}\right\vert_{s_j\in S}
\end{equation}
\end{small}
for $|S|>0$ and $\textsc{Save}@K = 0$ for $|S| = 0$. In Eq~\ref{eq: save-formula-positive-S}, $S$ is the set of SQL queries that are correctly predicted, given each prefix $q_j^{\text{pre}}$ of the complete question $q^{\text{cmp}}$. $s_j$ is one of the correctly predicted SQL queries in $S$. Function $\text{len}(\cdot)$ calculates the number of tokens.

For all the aforementioned metrics, we use the exact match result to judge the correctness.

\paragraph{Metrics' Ceiling Scores}
There exist certain examples in \datas where there are more than $5$ gold SQL queries so that models cannot achieve $100\%$ \textsc{Recall}. The maximum average \textsc{Recall}@5 and \textsc{Recall}@10 scores a model can achieve in \datas are $97\%$ to $99\%$ for each sub-task. For \textsc{MRR}, the maximum score a model can achieve will be $100\%$ across all sub-tasks if the first SQL prediction is among the gold SQL queries. For \textsc{Save}, the maximum score the model can achieve theoretically is 100\%, but in practice, it is impossible to reach $100\%$ because models cannot predict the correct corresponding SQL query with a zero-length prefix.

% Baseline Question split results for recall, mrr and save for 5
\begin{table*}[t]
\small
\centering
 \begin{tabular}{l   r  r  r  r  r  r  r  r  r  r  r  r  r  r  r  }
     \toprule
     \multirow{2}{*}{Model}  & \multicolumn{3}{c}{Advising} & \multicolumn{3}{c}{ATIS} & \multicolumn{3}{c}{GeoQuery} & \multicolumn{3}{c}{Scholar} & \multicolumn{3}{c}{Spider} \\
    & \multicolumn{1}{c}{R} & \multicolumn{1}{c}{M} & \multicolumn{1}{c}{S} 
     & \multicolumn{1}{c}{R} & \multicolumn{1}{c}{M} & \multicolumn{1}{c}{S}
      & \multicolumn{1}{c}{R} & \multicolumn{1}{c}{M} & \multicolumn{1}{c}{S}
       & \multicolumn{1}{c}{R} & \multicolumn{1}{c}{M} & \multicolumn{1}{c}{S}
        & \multicolumn{1}{c}{R} & \multicolumn{1}{c}{M} & \multicolumn{1}{c}{S} \\
    \midrule
    \textsc{Pf-S2S}   &17 & 14  & 8        &10  & 8   & 6    &24  & 20& 10    & 13 & 10  & 10 & \textbf{36} &  27  & \textbf{37}\\
     
    \textsc{ + Attn}  &16 & 14   & 8       &10 & 7 & 6      &27 &  21  &  11    & \textbf{14} & \textbf{12} &\textbf{11} & \textbf{36} & \textbf{28} & \textbf{37} \\
    
    \textsc{ + Copy} & \textbf{22}  &  \textbf{19} & \textbf{12}  &11 & 8 &  6    & 23 &  17&  11     & 10 & 7 &7  & 31 & 23 & 35\\
    
    \textsc{QAC-S2S}   & 8 & 7     &   6     & \textbf{13}  & \textbf{11} &  6    & \textbf{37} & \textbf{31} & \textbf{15}   & 11 &  9 & 9  &  - & - & -\\
    \multicolumn{16}{c}{}\\[-0.8em]
    \hdashline
    \multicolumn{16}{c}{}\\[-0.8em]the 
    \textsc{T5} &\textbf{45} & \textbf{39} & \textbf{29} &11 & 8 & 6  & \textbf{37} & \textbf{34} & 11 & \textbf{36} & \textbf{30} &\textbf{24} & 44 & 20 & 40\\
      \midrule
     \textsc{Emb-Rtr} &- & - & -     & -  & - & -           &22 & 18 & 8                   &2 & 1 & 1        & 10 & 5 & 11 \\
          
      \textsc{Cls-Rtr}  & \textbf{19} & \textbf{15} & \textbf{10}                   & \textbf{7} & \textbf{4} & \textbf{4}              & \textbf{26} & \textbf{21} & \textbf{10}             & \textbf{20} & \textbf{15} & \textbf{13}      & \textbf{60} & \textbf{52} & \textbf{57}\\
      \bottomrule
      
     \end{tabular}
\caption[Caption for LOF]{ \textsc{Recall}@5 (R), \textsc{MRR}@5 (M) and \textsc{Save}@5 (S) in percentage for each subtask in \datas on question split. We embolden the best scores for baselines other than T5 above and below the dashed line, respectively. For T5, we embolden its score if it is the highest among all the baselines. We use ``-'' to denote scores $<1\%$.}
 \label{tab: recall-5-generative-models}
 \end{table*}

%%%%%%%%%%%%%%%%%%%%%%%%%%%%%%%%%%%%%%%%%%%%%%%%%%%%%%%%%%%%%%%%%%%%%%%%%%%%%%%%%%%%%%%%%%%%%%%%%%%%%%%%%%
%%%%%%%%%%%%%%%%%%%%%%%%%%%%%%%%%%%%%%%%%%%%%%%%%%%%%%%%%%%%%%%%%%%%%%%%%%%%%%%%%%%%%%%%%%%%%%%%%%%%%%%%%%

% Models

\section{Baseline Models}

We regard \task task either as a SQL generation task or a SQL retrieval task. Thus, we experiment with baselines from both SQL generation models and retrieval models.

\paragraph{Generation-based Model}
We experiment with seq2seq models using a Bi-LSTM encoder and a LSTM decoder \cite{hochreiter1997long} directly on question prefixes to generate SQL. We name the seq2seq model  \textbf{\textsc{Pf-S2S}}. Seq2seq with attention (\textbf{\textsc{+ attn}}), as well as seq2seq with attention and copy mechanism (\textbf{\textsc{+ attn + copy}}) are also evaluated. \textbf{\textsc{T5}}~\cite{raffel2020exploring} is a pre-trained sequence-to-sequence model based on the Transformer architecture \cite{vaswani2017attention}. Following~\citet{shaw2021compositional,hazoom2021texttosql}, we use T5 as our baseline model. Unlike the traditional setting where one question has a single gold SQL, a question prefix can match multiple gold SQL queries in \task. Thus, our models optimize the sum of log-likelihood of all gold SQL queries following~\citet{jin2003learning}:

\begin{equation}
    \label{eq: loss-function}
   \mathcal{L}=\sum_{q^{\text{pre}}} \sum_{i=1}^{L} \log P(s^{\text{gold}}_i|q^{\text{pre}}) 
\end{equation}

We also use a two-stage model (\textbf{\textsc{QAC-S2S}}) that first uses the GPT2 language model \cite{radford2019language} to auto-complete question prefixes to form complete questions. Then seq2seq with attention and copy mechanism translates the complete question into SQL queries. The two stages are trained separately, and we select the top K SQL predictions during testing. 

\paragraph{Retrieval-based Model}

We finetune the RoBERTa \cite{liu2019roberta} model to generate the embeddings for both question prefixes and historical SQL queries (SQL queries in the training set). We use the dot-product of their embeddings to represent the similarity between question prefix and SQL query. Historical SQL queries with the K highest similarity scores will be retrieved for a question prefix during the inference. We use \textbf{\textsc{Emb-Rtr}} to denote this model. 

Additionally, we train a RoBERTa-based classification model (\textbf{\textsc{Cls-Rtr}}) to distinguish relevant SQL queries from irrelevant SQL queries. The model predicts whether the SQL matches the intention of the question prefix directly. During inference, we rank all historical SQL queries and retrieve top K SQL queries based on the predicted probability by \textbf{\textsc{Cls-Rtr}}.

%%%%%%%%%%%%%%%%%%%%%%%%%%%%%%%%%%%%%%%%%%%%%%%%%%%%%%%%%%%%%%%%%%%%%%%%%%%%%%%%%%%%%%%%%%%%%%%%%%%%%%%%%%
%%%%%%%%%%%%%%%%%%%%%%%%%%%%%%%%%%%%%%%%%%%%%%%%%%%%%%%%%%%%%%%%%%%%%%%%%%%%%%%%%%%%%%%%%%%%%%%%%%%%%%%%%%

% Experiment Setup
% Already merged into the previous sections

%%%%%%%%%%%%%%%%%%%%%%%%%%%%%%%%%%%%%%%%%%%%%%%%%%%%%%%%%%%%%%%%%%%%%%%%%%%%%%%%%%%%%%%%%%%%%%%%%%%%%%%%%%
%%%%%%%%%%%%%%%%%%%%%%%%%%%%%%%%%%%%%%%%%%%%%%%%%%%%%%%%%%%%%%%%%%%%%%%%%%%%%%%%%%%%%%%%%%%%%%%%%%%%%%%%%%

% Results and Analysis

 \begin{table}[t]
 \small
 \setlength{\tabcolsep}{5pt}
 \centering

    \begin{tabular}{l   r  r  r  r  r  r  r  r  r }
     \toprule
     \multirow{2}{*}{Model}  & \multicolumn{3}{c}{ATIS} & \multicolumn{3}{c}{GeoQuery} & \multicolumn{3}{c}{Spider} \\
    & \multicolumn{1}{c}{R} & \multicolumn{1}{c}{M} & \multicolumn{1}{c}{S} 
     & \multicolumn{1}{c}{R} & \multicolumn{1}{c}{M} & \multicolumn{1}{c}{S}
      & \multicolumn{1}{c}{R} & \multicolumn{1}{c}{M} & \multicolumn{1}{c}{S}\\
        \midrule
    
    \textsc{Pf-S2S}  & \textbf{3} & 1 &  1               & 7 &3 &  2                 & 2 & 1 & 3\\
     
    \textsc{ + Attn}  & \textbf{3} & 1 & 1            & \textbf{8} & \textbf{4} & \textbf{3}         & \textbf{3} & 1 & \textbf{4}\\
    
    \textsc{ + Copy}  & 2 & 1 &  1                     & 5 &2 & 2                  & 2 & 1 & 2 \\
    
    \textsc{QAC-S2S}  & 2 & 1 &  1                      & 7 & \textbf{4} &  \textbf{3}                 &  - & - & -\\
    \multicolumn{6}{c}{}\\[-0.8em]
    \hdashline
    \multicolumn{6}{c}{}\\[-0.8em]
    \textsc{T5} & 2 & 1 & 1                              & \textbf{15} & \textbf{11} & \textbf{5}           & \textbf{9} & \textbf{6} & \textbf{8}\\

    \bottomrule
     
     \end{tabular}

 \caption[Caption for LOF]{\textsc{Recall}@5 (R), \textsc{MRR}@5 (M) and \textsc{Save}@5 (S) in percentage generative models on SQL query split. We omit the results for sub-tasks Advising and Scholar because models perform $0-1\%$ for all metrics. We use ``-'' to denote scores $<1\%$.}
 \label{tab: scores-query-split-generative}
 \end{table}

\section{Baseline Results and Analysis}
\label{sec:results}

\paragraph{Overall Results.}

Table~\ref{tab: recall-5-generative-models} shows scores in terms of the three metrics on \datas. Models that achieve good \textsc{Recall} scores also achieve good \textsc{MRR} and \textsc{Save} scores, which indicates that models predicting the most number of correct gold SQL queries also rank the correct SQL queries higher and save more user efforts.

We do not report scores for retrieval models on SQL query split because retrieval models fail to retrieve unseen SQL queries. But in question split where the same SQL query might appear in training, \textsc{Cls-Rtr} achieves a \textsc{Recall}@5 of $60\%$, outperforming generative models on sub-task Spider (the highest \textsc{Recall}@5 for generative models is $44\%$). Although \textsc{Cls-Rtr} outperforms \textsc{Emb-Rtr} on all the sub-tasks, \textbf{\textsc{Cls-Rtr}} requires much more running time than \textbf{\textsc{Emb-Rtr}} because it needs to run the RoBERTa classification model on all historical SQL queries with the given question prefix. In contrast, \textbf{\textsc{Emb-Rtr}} can calculate and cache historical SQL embeddings in advance.

\textsc{QAC-S2S} achieves a good \textsc{Recall}@5 ($37\%$) on sub-task GeoQuery (same as \textsc{Recall}@5 for \textsc{T5}). However, \textsc{QAC-S2S} performs poorly on sub-task Spider ($<1\%$ \textsc{Recall}@5) because the second stage is doing the original text-to-SQL on complete questions, and seq2seq models perform poorly on the original Spider ($16.0\%$ reported by \cite{yu-etal-2018-spider} compared to $71\%$ on the original GeoQuery dataset reported by~\cite{finegan2018improving}). The cascade of errors in the two stages result in the poor performance of \textsc{QAC-S2S} on sub-task Spider.

\textsc{T5} is a strong baseline as it achieves the highest \textsc{Recall}@5 scores on sub-tasks Advising and Scholar ($45\%$ and $36\%$, respectively). In terms of \textsc{Recall}@5 scores, \textsc{T5} outperforms variants of \textsc{PF-S2S} ($22\%$ on Advising, $14\%$ on Scholar) and \textsc{QAC-S2S} ($8\%$ on Advising, $11\%$ on Scholar) by a large margin on Advising and Scholar. So are the cases for \textsc{MRR} and \textsc{Save} (For \textsc{MRR}@5 and \textsc{Save}@5, T5 achieves $39\%$ and $29\%$ on Advising while the second-highest are $19\%$ and $12\%$, respectively). We use T5 as a strong baseline for our later curriculum learning setting because the performance of \textsc{PF-S2S} is comparable to \textsc{QAC-S2S} in most cases, and there is no clear indication that auto-completing prefix will benefit our task. 

\begin{figure*}[ht!]
    \centering
   \begin{subfigure}[c]{0.3\textwidth}
        \includegraphics[width=\textwidth]{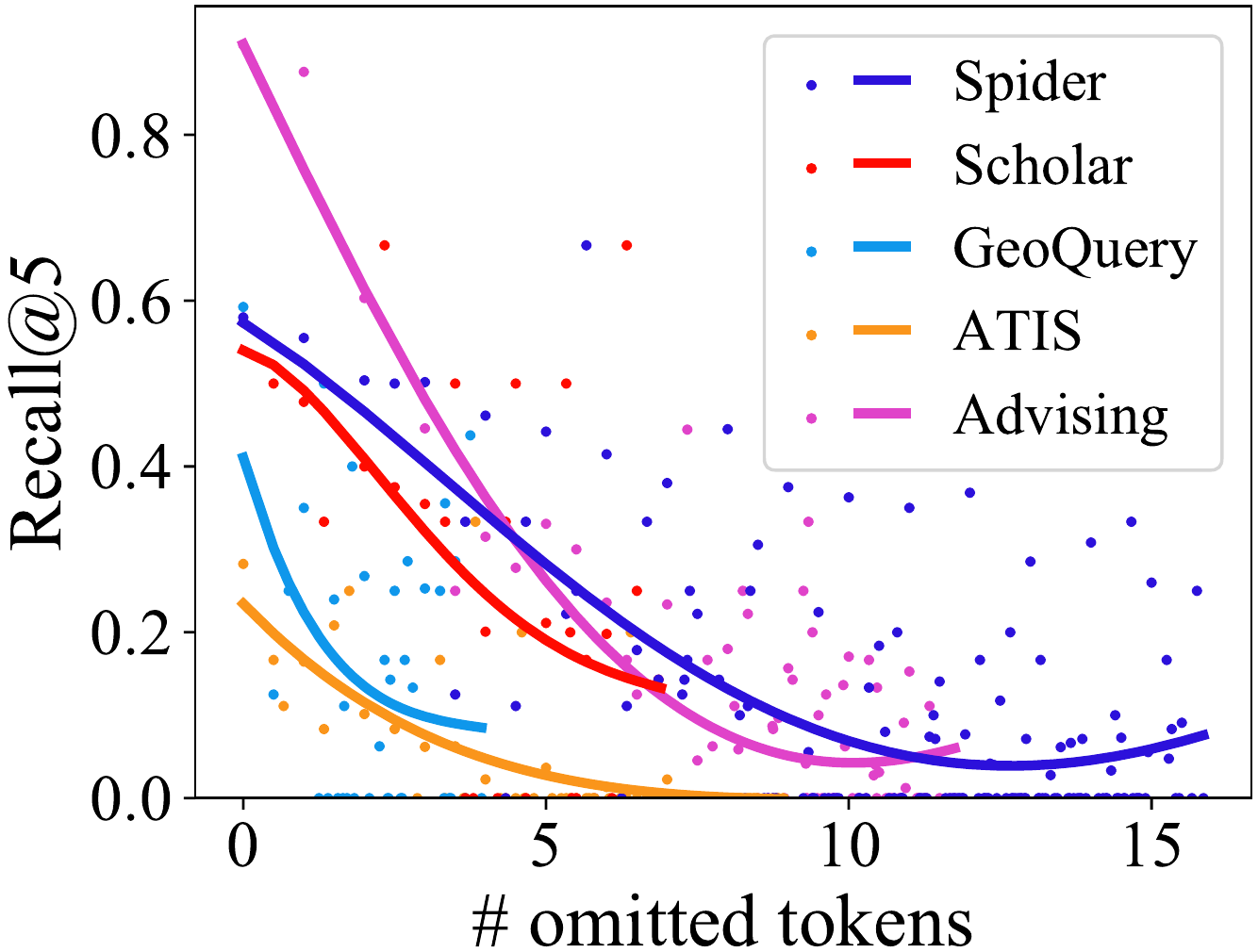}\caption{}
    \end{subfigure}\hspace{5mm}
    ~ 
    \begin{subfigure}[c]{0.3\textwidth}
    \includegraphics[width=\textwidth]{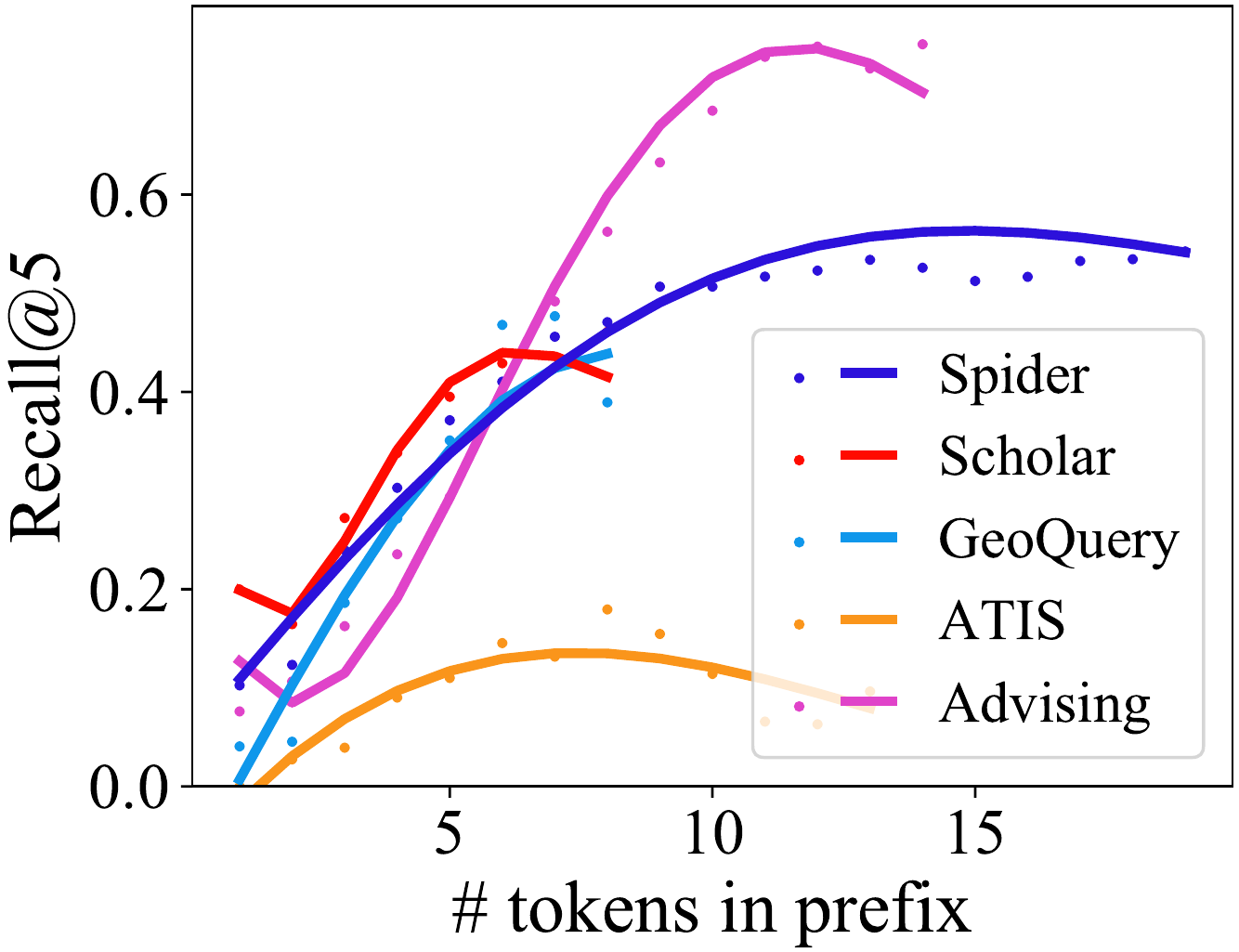}\caption{}
    \end{subfigure}
     ~ 
    \begin{subfigure}[c]{0.3\textwidth}
    \includegraphics[width=\textwidth]{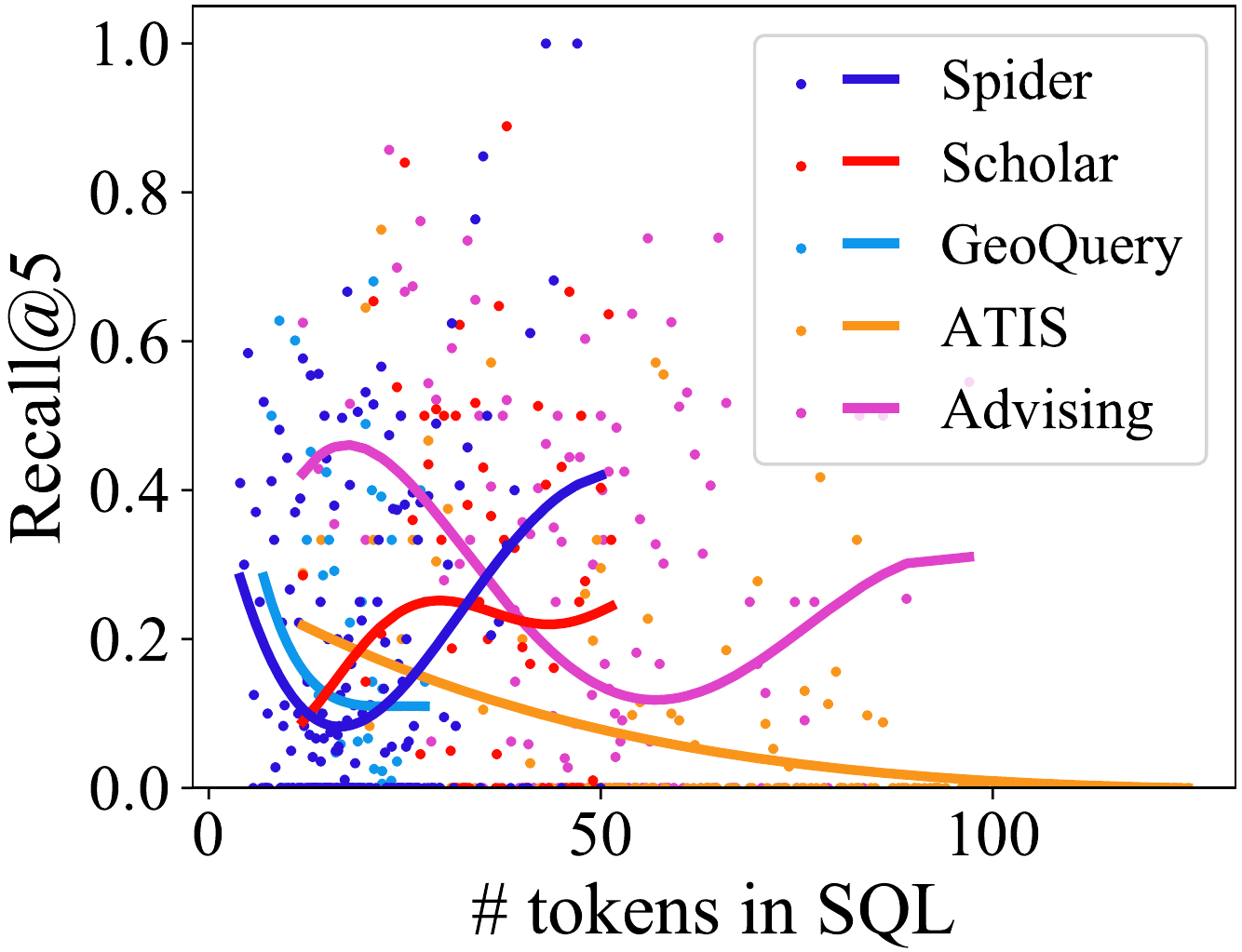}\caption{}
    \end{subfigure}
    \caption{T5's (a) \textsc{Recall}@5 v.s. number of omitted tokens, (b) \textsc{Recall}@5 v.s. prefix length (number of tokens in prefix), (c) \textsc{Recall}@5 v.s. SQL length (number of tokens in SQL) on question split for each sub-task in \datas. We plot \textsc{Recall}@5 corresponding to $\ge 50$ and $\ge 100$ examples for sub-tasks other than Spider and Spider, respectively. \textsc{Recall}@5 is negatively correlated with number of omitted tokens and possesses no monotonic relationships with either prefix length or SQL length.}
    \label{fig: perform-different-setting}
\end{figure*}

\paragraph{Generalization to Unseen SQL queries} As discussed in Section~\ref{subsec: dataset-construction}, SQL query split is a challenging setting because of the unseen SQL templates in the testing set. Models perform poorly on SQL query split for Advising and Scholar in \datas and only achieve $0-1\%$ for all the \textsc{Recall}@5, \textsc{MRR}@5 and \textsc{Save}@5. Even on SQL query split for ATIS, Scholar, and Spider in \datas where models achieve non-zero \textsc{Recall}@5 scores (Table~\ref{tab: scores-query-split-generative}), the scores are overshadowed by their corresponding scores on question split. 

Although generative models perform poorly on SQL query split, their scores on GeoQuery ($15\%$ \textsc{Recall}@5 for T5) and Spider ($9\%$ \textsc{Recall}@9 for T5) in \datas demonstrate their ability to generalize to unseen SQL queries. As shown in Table~\ref{tab: scores-query-split-generative}, T5 achieves \textsc{Save}@5 scores of $5\%$ and $8\%$ for GeoQuery and Spider, respectively. This indicates that models can still save some users' effort even if the user requests are different from the existing ones.

\begin{figure}[t!]
    \centering
    \includegraphics[width=0.45\textwidth]{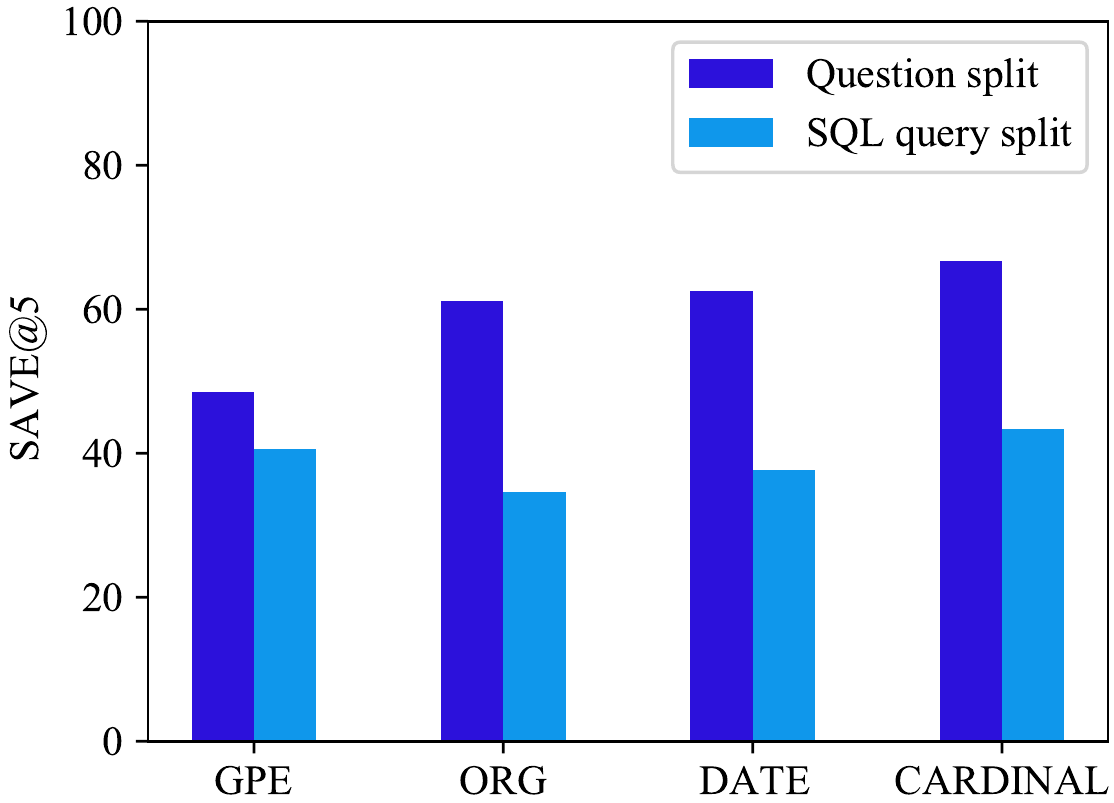}
    \caption{\textsc{Save}@5 v.s. omitted entity types for T5 for the top $4$ most popular omitted entity types. T5 achieves similar \textsc{Save}@5 of around $50\%$ to $60\%$ on GPE, ORG, DATE, and CARDINAL on question split, and $35\%$ to $40\%$ on SQL query split. }
    \label{fig: baseline-ent-type-SAVE}
\end{figure}

\paragraph{Performance Analysis}
Figure~\ref{fig: baseline-ent-type-SAVE} shows T5's \textsc{Save}@5 v.s. the omitted entity types. T5 achieves similar \textsc{Save}@5 of around $50\%$ to $60\%$ on GPE, ORG, DATE, and CARDINAL on question split, and $35\%$ to $40\%$ on SQL query split. In other words, if users intend to ask questions including GPE (geopolitical entity), the model can auto-complete that entity for around $50\%$ of the time if it sees the SQL template before.

In Figure~\ref{fig: perform-different-setting}, T5's \textsc{Recall}@5 scores are negatively correlated with the number of omitted tokens but possess no monotonic relationships with either the prefix length (number of tokens in the prefix) or SQL length (number of tokens in SQL queries). This differs \task task from the original complete text-to-SQL task, as short SQL queries are considered easier than longer ones in the original text-to-SQL task~\cite{finegan2018improving}. As discussed in Section~\ref{subsec: dataset-analysis}, the number of omitted tokens possesses a positive linear relationship with the number of omitted entities as shown in Figure~\ref{subfig:question_omit_len_ent_ratio}. The hardness of the task is negatively correlated with the number of omitted tokens or the number of omitted entities as well.

%%%%%%%%%%%%%%%%%%%%%%%%%%%%%%%%%%%%%%%%%%%%%%%%%%%%%%%%%%%%%%%%%%%%%%%%%%%%%%%%%%%%%%%%%%%%%%%%%%%%%%%%%%
%%%%%%%%%%%%%%%%%%%%%%%%%%%%%%%%%%%%%%%%%%%%%%%%%%%%%%%%%%%%%%%%%%%%%%%%%%%%%%%%%%%%%%%%%%%%%%%%%%%%%%%%%%

% curriculum learning

\section{Curriculum Learning} 

\begin{figure}[t!]
    \centering
    \includegraphics[width=0.48\textwidth]{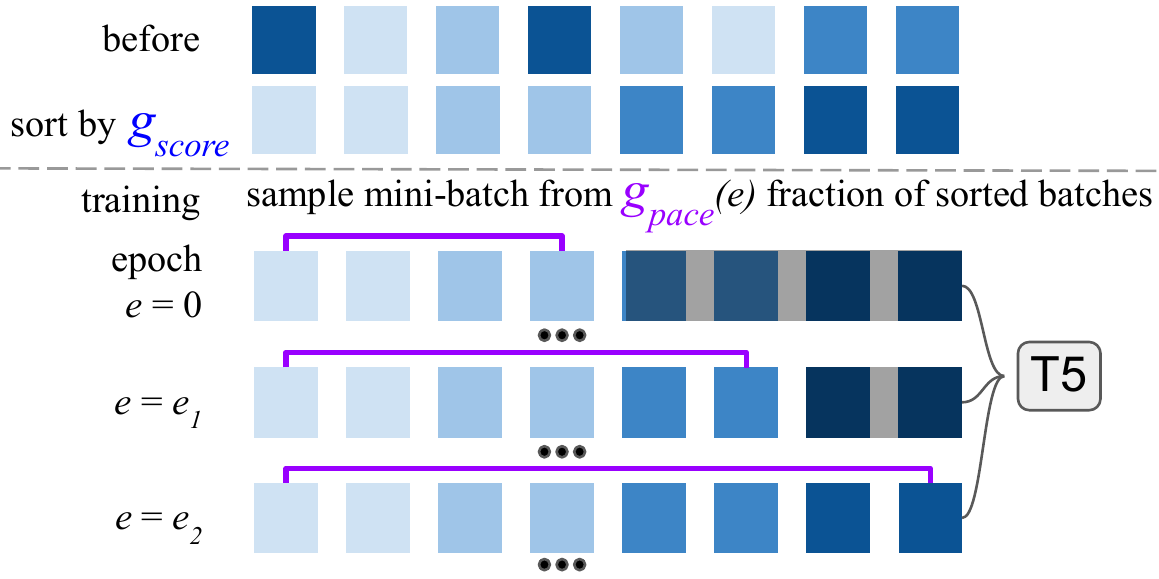}
    \caption{Setups for curriculum learning with the T5 model. $\textsl{g}_{\text{score}}$ defines the difficulties of each batch. We use different shades to express the difficulties of each batch, with darker blue indicating a more difficult batch while lighter indicate an easier batch. For each epoch, we sample batches by $\textsl{g}_{\text{pace}}$ and only feed those batches to T5.}
    \label{fig: T5-curriculum}
\end{figure}

\begin{table*}[t!]
 \small
 \centering
 
  \begin{tabular}{l   r  r  r  r  r  r  r  r  r  r  r  r  r  r  r  }
     \toprule
     \multirow{2}{*}{Model}  & \multicolumn{3}{c}{Advising} & \multicolumn{3}{c}{ATIS} & \multicolumn{3}{c}{GeoQuery} & \multicolumn{3}{c}{Scholar} & \multicolumn{3}{c}{Spider} \\
    & \multicolumn{1}{c}{R} & \multicolumn{1}{c}{M} & \multicolumn{1}{c}{S} 
     & \multicolumn{1}{c}{R} & \multicolumn{1}{c}{M} & \multicolumn{1}{c}{S}
      & \multicolumn{1}{c}{R} & \multicolumn{1}{c}{M} & \multicolumn{1}{c}{S}
       & \multicolumn{1}{c}{R} & \multicolumn{1}{c}{M} & \multicolumn{1}{c}{S}
        & \multicolumn{1}{c}{R} & \multicolumn{1}{c}{M} & \multicolumn{1}{c}{S} \\
    \midrule
    
      \textsc{T5} & 45.0 & 38.8  & 28.8                 & 10.7 & 7.9 & 5.7               & 36.8 & 33.5& 10.8                                 & 35.8 & 30.0 & 24.4                        & 43.7 &19.7 & 39.9\\
          
      \textsc{T5\SPSB{+CL}{(ours)}}  & 42.0 & 37.9 & 27.2                 & \textbf{15.4} & \textbf{12.6} & \textbf{7.7}                     & \textbf{42.7} & \textbf{40.0}& \textbf{14.7}            & \textbf{37.2} & \textbf{31.9}& 23.2          & \textbf{45.0}  & \textbf{36.1}& \textbf{41.4} \\
      
    \multicolumn{6}{c}{}\\[-0.8em]
    \hdashline
    \multicolumn{6}{c}{}\\[-0.8em]
    
       \textsc{T5} & 0.7 & 0.3 & 0.5             & 1.7 & 1.1 & 0.8            & 14.7 & 10.6 & 5.2         & 0.3 & 0.1 & 0.4                & 9.0 & 6.1 & 7.7  \\
          
      \textsc{T5\SPSB{+CL}{(ours)}}  & \textbf{2.8}& \textbf{2.0} & \textbf{2.0}                    & \textbf{5.1} & \textbf{3.4}  & \textbf{2.6}                             & \textbf{24.0}   & \textbf{17.6} & \textbf{6.4}                         & \textbf{1.1} & \textbf{0.6}& 0.4                       & 7.8 & 5.0  & 7.0\\

      \bottomrule
     
     \end{tabular}

 \caption[Caption for LOF]{\textsc{Recall}@5 (R), \textsc{MRR}@5 (M) and \textsc{Save}@5 (S) in percentage for T5 and T5 together with curriculum learning (our proposed method) for question split (upper half) and SQL query split (lower part). We embolden score improvement after incorporating the curriculum learning.}
 \label{tab: T5-CL-recall-result-question-query-split}
 \end{table*}

Based on our discovery of the relationship between \textsc{Recall} and the number of omitted tokens, we propose the use of curriculum learning~\cite{bengio2009curriculum} to improve the T5 performance. The setups are shown in Figure~\ref{fig: T5-curriculum}. We propose the scoring function as:
\begin{equation}
\small
\label{eq: difficulty_score}
        \textsl{g}_{\text{score}} = \text{len}(q^{\text{pre}}) - \text{len}(q^{\text{cmp}})
\end{equation}
to score the difficulty of each example (based on Eq~\ref{eq: loss-function}, we use $q^{\text{pre}}$ for each SQL query in $S^{\text{gold}}$ as a single data-point during training. Thus, we use the corresponding $q^{\text{cmp}}$ for every $s\in S^{\text{gold}}$ for $\textsl{g}_{\text{score}}$).
For pacing function, we choose
\begin{equation}
\small
\label{eq: pacing-function}
    \textsl{g}_{\text{pace}}(e) = \begin{cases}
        \left.B\right\vert_{d_{\text{min}} \leq d \leq d_{\text{min}} + m} & e = 0\\
        \left.B\right\vert_{d_{\text{min}} + m \leq d \leq d_{\text{min}} + m + \lambda e} & 1 \leq e \leq \frac{\tau}{\lambda} \\
        \left.B\right\vert_{d_{\text{min}} \leq d \leq d_{\text{max}}} & \frac{\tau}{\lambda} \leq e
    \end{cases}
\end{equation}
where $\tau = d_{\text{max}} - d_{\text{min}} - m $. $B$ represents the set of mini-batches sorted by $\textsl{g}_{\text{score}}$, $d$ represents the score calculated by $\textsl{g}_{\text{score}}$. For epoch $e = 0$, $\textsl{g}_{\text{pace}}$ selects mini-batches with difficulties from $d_{\text{min}}$ $d_{\text{min}} + m$. Then for each epoch $e$, we will include mini-batches with an increasing difficulty of $\lambda e$ until the difficulty reaches $d_{\text{max}}$. After that, we will always include all the batches in the following epochs. 

We use T5 for curriculum learning (\textbf{\textsc{T5+CL}}). For the hyper-parameters in Eq~\ref{eq: pacing-function}, we tune $\lambda$ in $\{1, 2\}$ and $m$ in $\{1, 2, 3\}$ for sub-task GeoQuery on question split in \datas, and find that $\lambda=1, m=2$ works best for the curriculum learning. We then set $\lambda=1, m=2$ for both splits for all domains in \datas.

\subsection{Results and Analysis}
Table~\ref{tab: T5-CL-recall-result-question-query-split} shows the comparison of \textsc{Recall}@5, \textsc{MRR}@5 and \textsc{Save}@5 scores between the original T5 and T5 with curriculum learning. In \datas, T5 with curriculum learning improves \textsc{Recall}@5 on ATIS, GeoQuery, Scholar and Spider for question split, on Adivising, ATIS, GeoQuery and Scholar for SQL query split. It improves \textsc{Recall}@5 for as much as $9\%$ on SQL query split for ATIS (\textsc{T5+CL} achieves $24.0\%$ while \textsc{T5} achieves $14.7\%$); \textsc{MRR}@5 for as much as $16\%$ on question split for Spider (\textsc{T5+CL} achieves $36.1\%$ while \textsc{T5} achieves $19.7\%$); \textsc{Save}@5 for as much as $3.9\%$ on question split for GeoQuery (\textsc{T5+CL} achieves $14.7\%$ while \textsc{T5} achieves $10.8\%$).

\begin{figure}[t!]
    \centering
    \includegraphics[width=0.45\textwidth]{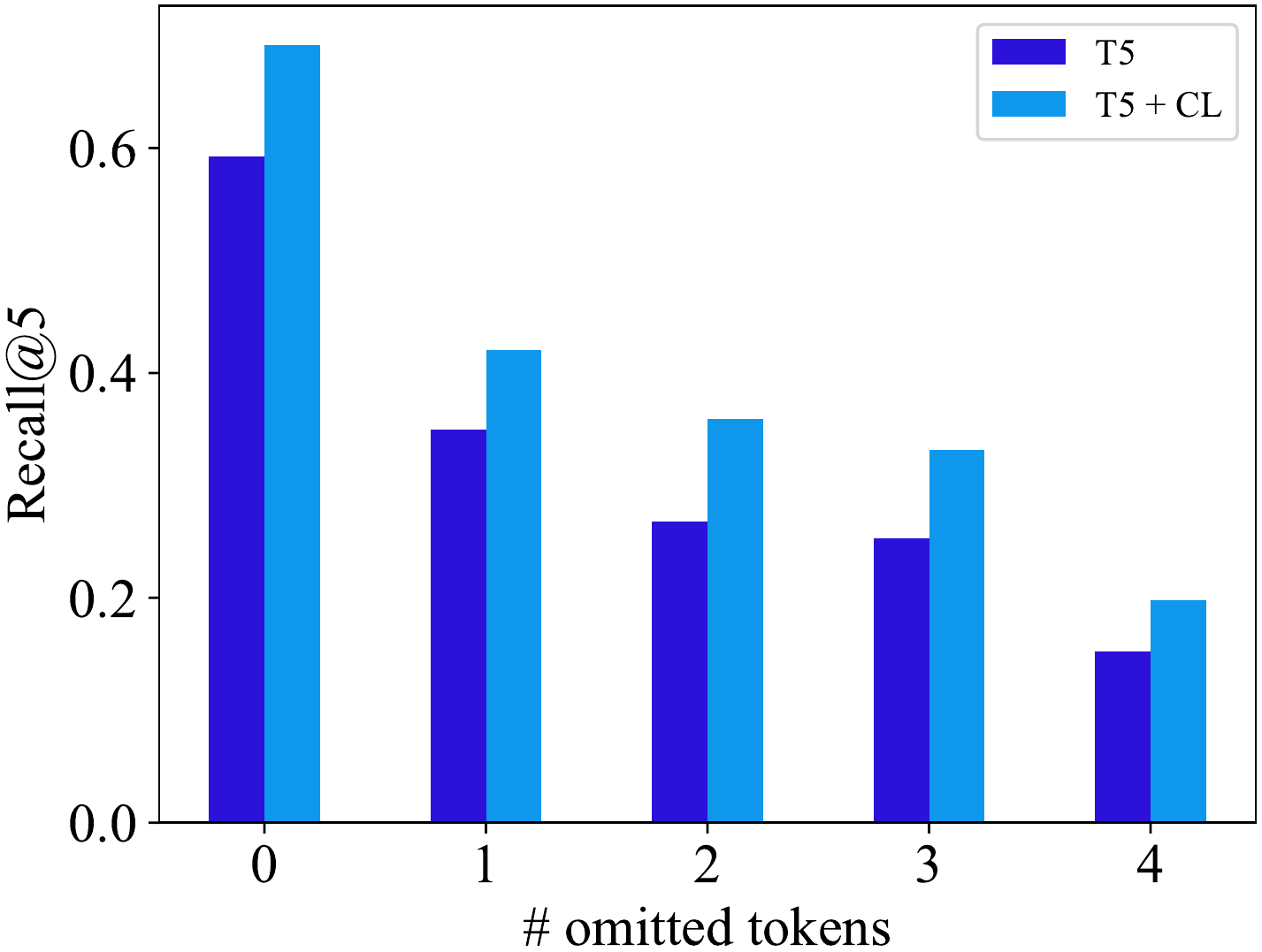}
    \caption{\textsc{Recall}@5 score comparison for different omitted number of tokens on sub-task GeoQuery. We report \textsc{Recall}@5 for number of omitted tokens corresponding to $\ge 50$ examples.}
    \label{fig: CL-recall-missing-len}
\end{figure}

\begin{figure}[t!]
    \centering
    \includegraphics[width=0.45\textwidth]{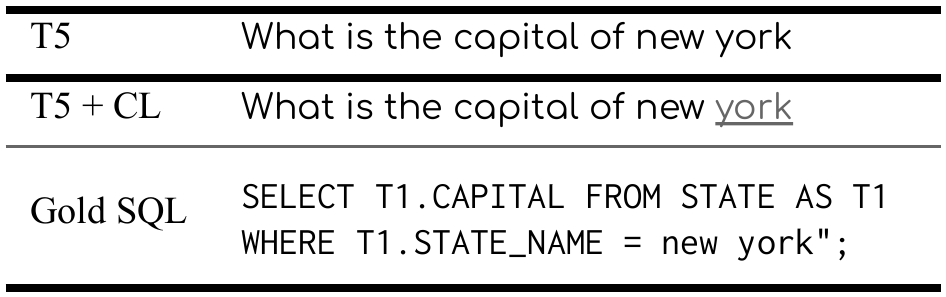}
    \caption{Prefixes with the least number of tokens where both models can predict the correct gold SQL in sub-task GeoQuery. \textsc{T5+CL} correctly predicts the gold SQL with one less token (\underline{underlined} text) in prefix than the original \textsc{T5}.}
    \label{fig: CL-prelen-improve-example}
\end{figure}

Figure~\ref{fig: CL-recall-missing-len} shows comparison of \textsc{Recall}@5 for \textsc{T5} and \textsc{T5 + CL} for sub-task GeoQuery on question split. By involving the curriculum, the model achieves better \textsc{Recall}@5 scores on different levels of difficulties for \task. Although \textsc{T5+CL} improves performances for various sub-tasks, for sub-task advising on question split and Spider on SQL query split, we find that \textsc{T5+CL} performs worse than \textsc{T5}, indicating sub-tasks Advising and Spider might have their own challenges.

Figure~\ref{fig: CL-prelen-improve-example} shows an example in sub-task GeoQuery that is correctly predicted by \textsc{T5+CL} with one less token than T5. \textsc{T5+CL} successfully predict ``York'' as the completion for the prefix ``what is the capital of New''. And this does not impair \textsc{T5+CL}'s performance on prefixes that end with ``New''. On examples involving ``New Jersey'', ``New Mexico'' (examples which use ``Jersy'' or ``Mexico'' as the completion for prefixes that end with ``New''), \textsc{T5+CL} performs as good as the original \textsc{T5}.

%%%%%%%%%%%%%%%%%%%%%%%%%%%%%%%%%%%%%%%%%%%%%%%%%%%%%%%%%%%%%%%%%%%%%%%%%%%%%%%%%%%%%%%%%%%%%%%%%%%%%%%%%%
%%%%%%%%%%%%%%%%%%%%%%%%%%%%%%%%%%%%%%%%%%%%%%%%%%%%%%%%%%%%%%%%%%%%%%%%%%%%%%%%%%%%%%%%%%%%%%%%%%%%%%%%%%

% Conclusion

\section{Conclusion}

In this work, we propose \task and construct a benchmark \datas, making the first step to build a more user-friendly NLIDB system. To better evaluate models' performance, apart from \textsc{Recall} and \textsc{MRR}, we propose our own metric \textsc{Save} which measures how much user effort can be saved. Experiments show that \datas is challenging even for strong baseline models such as \textsc{T5}. Analysis shows that different from the original text-to-SQL, the difficulty of \task is related to the number of omitted tokens. Based on this discovery, we incorporate curriculum learning of feeding examples with an increasing number of omitted tokens. This improves scores on various sub-tasks in \datas and metrics, and by as much as $9\%$ \textsc{Recall} score and $3.9\%$ \textsc{Save} score on sub-task GeoQuery. However, even with curriculum learning, there is a large room for improvement for current models, indicating the necessity of future research.

%%%%%%%%%%%%%%%%%%%%%%%%%%%%%%%%%%%%%%%%%%%%%%%%%%%%%%%%%%%%%%%%%%%%%%%%%%%%%%%%%%%%%%%%%%%%%%%%%%%%%%%%%%
%%%%%%%%%%%%%%%%%%%%%%%%%%%%%%%%%%%%%%%%%%%%%%%%%%%%%%%%%%%%%%%%%%%%%%%%%%%%%%%%%%%%%%%%%%%%%%%%%%%%%%%%%%

% Acknowledgment

\section*{Acknowledgment}
We would like to thank Yulong Chen for his valuable advice on the abstract and introduction. We would also like to thank Chao Chen, Yiqun Yao, Yin Lin, Lingyi Jin, Laura Biester, Andrew Lee and Nan Zhang for proofreading this paper.
This work is in part supported by an Amazon Research Award. 
% proof read for this paper.

%%%%%%%%%%%%%%%%%%%%%%%%%%%%%%%%%%%%%%%%%%%%%%%%%%%%%%%%%%%%%%%%%%%%%%%%%%%%%%%%%%%%%%%%%%%%%%%%%%%%%%%%%%
%%%%%%%%%%%%%%%%%%%%%%%%%%%%%%%%%%%%%%%%%%%%%%%%%%%%%%%%%%%%%%%%%%%%%%%%%%%%%%%%%%%%%%%%%%%%%%%%%%%%%%%%%%
% References
% Reference page limit: 2 pages
% placed to the very end of the paper

\bibliography{anthology, custom}
\bibliographystyle{acl_natbib}

\clearpage
\newpage
\clearpage
\newpage
\appendix

%%%%%%%%%%%%%%%%%%%%%%%%%%%%%%%%%%%%%%%%%%%%%%%%%%%%%%%%%%%%%%%%%%%%%%%%%%%%%%%%%%%%%%%%%%%%%%%%%%%%%%%%%%
%%%%%%%%%%%%%%%%%%%%%%%%%%%%%%%%%%%%%%%%%%%%%%%%%%%%%%%%%%%%%%%%%%%%%%%%%%%%%%%%%%%%%%%%%%%%%%%%%%%%%%%%%%

% Appendix

\section{Appendices}
\label{sec:appendix}

\subsection{Canonical Question Generation}
\label{sec:appendix-canonical}
We collect 118 SQL queries and question templates. Table \ref{tab:canonical-question} lists top 15 examples of SQL and question templates. It is worth mentioning that the listed top 15 rules can cover $80\%$ of the question and SQL pairs from the original Spider dataset, which implies that we only need to collect a certain amount of question and SQL templates, and those templates can still cover a decent amount of possible SQL queries that users want. 

\citet{finegan2018improving} classifies question and SQL templates from the original ATIS, Advising, GeoQuery, and Scholar\footnote{https://github.com/jkkummerfeld/text2sql-data/tree/master/data}. 

\begin{table*}[htbp]
\begin{center}
\begin{tabular}{ |p{0.2cm}|c|p{11.5cm}| } 
\hline
\# & & \\
\hline
\multirow{2}{2em}{1} & SQL-template & \{SELECT0\} \{FROM\} WHERE \{COLUMN0\} \{OP0\} \{VALUE0\} \\
\cline{2-3}
& Question-template & Find \{SELECT0\}  whose \{COLUMN0\} \{OP0\} \{VALUE0\} . \\
\hline
\multirow{2}{2em}{2} & SQL-template & \{SELECT0\} \{FROM\} \\
\cline{2-3}
& Question-template & Find \{SELECT0\} . \\
\hline
\multirow{2}{2em}{3} & SQL-template & \{SELECT0\} \{FROM\} WHERE \{COLUMN0\} \{OP0\} \{VALUE0\} AND \{COLUMN1\} \{OP1\} \{VALUE1\} \\
\cline{2-3}
& Question-template & Find \{SELECT0\}  whose \{COLUMN0\} \{OP0\} \{VALUE0\} and \{COLUMN1\} \{OP1\} \{VALUE1\} . \\
\hline
\multirow{2}{2em}{4} & SQL-template & \{SELECT0\} \{FROM\} GROUP BY \{COLUMN0\} \\
\cline{2-3}
& Question-template & For each \{COLUMN0\} what is \{SELECT0\} . \\
\hline
\multirow{2}{2em}{5} & SQL-template & \{SELECT0\} \{FROM\} GROUP BY \{COLUMN0\} ORDER BY \{AGG0\} ( * ) \{SC0\} LIMIT \{VALUE0\} \\
\cline{2-3}
& Question-template & What \{COLUMN0\} are the top \{VALUE0\} \{AGG0\} \{SC0\} . List \{SELECT0\} . \\
\hline
\multirow{2}{2em}{6} & SQL-template & \{SELECT0\} \{FROM\} ORDER BY \{COLUMN0\} \{SC0\} LIMIT \{VALUE0\} \\
\cline{2-3}
& Question-template & List \{SELECT0\} that are the top \{VALUE0\} ranked by \{COLUMN0\} \{SC0\} . \\
\hline
\multirow{2}{2em}{7} & SQL-template & \{SELECT0\} \{FROM\} GROUP BY \{COLUMN0\} HAVING \{AGG0\} ( * ) \{OP0\} \{VALUE0\} \\
\cline{2-3}
& Question-template & For \{COLUMN0\} whose \{AGG0\} \{OP0\} \{VALUE0\} . List \{SELECT0\} . \\
\hline
\multirow{2}{2em}{8} & SQL-template & \{SELECT0\} \{FROM\} WHERE \{COLUMN0\} \{OP0\} ( \{SELECT1\} \{FROM\} ) \\
\cline{2-3}
& Question-template & Find \{SELECT0\} that have \{COLUMN0\} \{OP0\} \{SELECT1\} . \\
\hline
\multirow{2}{2em}{9} & SQL-template & \{SELECT0\} \{FROM\} ORDER BY \{COLUMN0\} \{SC0\} \\
\cline{2-3}
& Question-template & Show \{SELECT0\} with \{COLUMN0\} \{SC0\}  \\
\hline
\multirow{2}{2em}{10} & SQL-template & \{SELECT0\} \{FROM\} WHERE \{COLUMN0\} \{OP0\} \{VALUE0\} INTERSECT \{SELECT1\} \{FROM\} WHERE \{COLUMN1\} \{OP1\} \{VALUE1\} \\
\cline{2-3}
& Question-template & Show \{SELECT0\} ( or \{SELECT1\} ) that have both \{COLUMN0\} \{OP0\} \{VALUE0\} and \{COLUMN1\} \{OP1\} \{VALUE1\} . \\
\hline
\multirow{2}{2em}{11} & SQL-template & \{SELECT0\} \{FROM\} WHERE \{COLUMN0\} \{OP0\} \{VALUE0\} OR \{COLUMN1\} \{OP1\} \{VALUE1\} \\
\cline{2-3}
& Question-template & Find \{SELECT0\}  whose \{COLUMN0\} \{OP0\} \{VALUE0\} or \{COLUMN1\} \{OP1\} \{VALUE1\} . \\
\hline
\multirow{2}{2em}{12} & SQL-template & \{SELECT0\} \{FROM\} WHERE \{COLUMN0\} NOT IN ( \{SELECT1\} \{FROM\} ) \\
\cline{2-3}
& Question-template & Find \{SELECT0\}  whose \{COLUMN0\} is not in \{SELECT1\} . \\
\hline
\multirow{2}{2em}{13} & SQL-template & \{SELECT0\} \{FROM\} ORDER BY \{COLUMN0\} \\
\cline{2-3}
& Question-template & What are \{SELECT0\}  ordered by \{COLUMN0\} . \\
\hline
\multirow{2}{2em}{14} & SQL-template & \{SELECT0\} \{FROM\} WHERE \{COLUMN0\} \{OP0\} \{VALUE0\} AND \{COLUMN1\} \{OP1\} \{VALUE1\} AND \{COLUMN2\} \{OP2\} \{VALUE2\} \\
\cline{2-3}
& Question-template & Find \{SELECT0\}  whose \{COLUMN0\} \{OP0\} \{VALUE0\} and \{COLUMN1\} \{OP1\} \{VALUE1\} and \{COLUMN2\} \{OP2\} \{VALUE2\} . \\
\hline
\multirow{2}{2em}{15} & SQL-template & \{SELECT0\} \{FROM\} WHERE \{COLUMN0\} \{OP0\} \{VALUE0\} GROUP BY \{COLUMN1\} \\
\cline{2-3}
& Question-template & For each \{COLUMN1\} whose \{COLUMN0\} \{OP0\} \{VALUE0\}  list \{SELECT0\} . \\
\hline
\end{tabular}
\caption{Top 15 SQL and canonical question templates.}
\label{tab:canonical-question}
\end{center}
\end{table*}

\begin{figure*}[ht!]
    \centering
   \begin{subfigure}[c]{0.4\textwidth}
        \includegraphics[width=\textwidth]{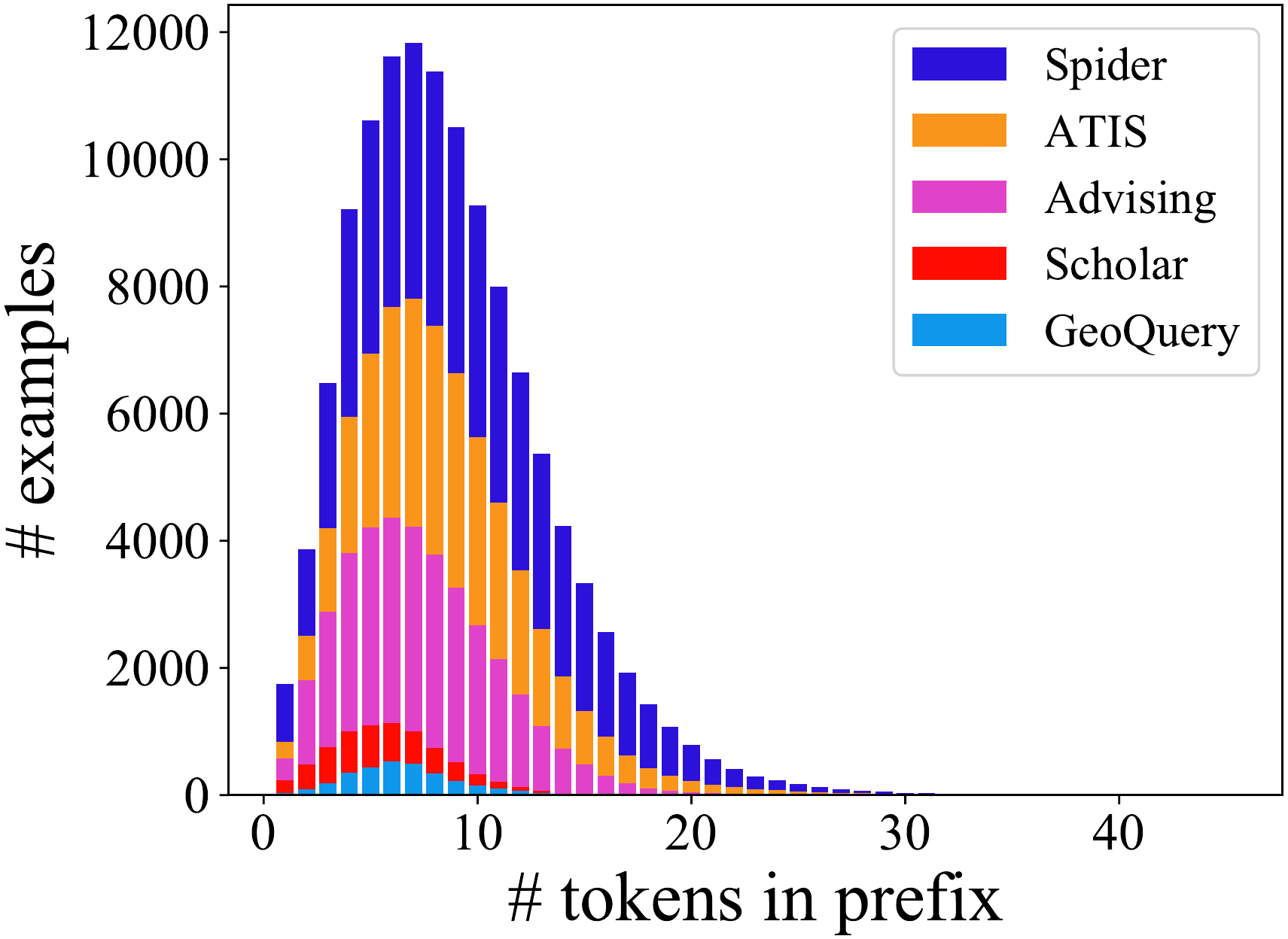}\caption{}
        \label{subfig:question_prefix_len_distribution}
    \end{subfigure}\hspace{5mm}
    ~ 
    \begin{subfigure}[c]{0.4\textwidth}
    \includegraphics[width=\textwidth]{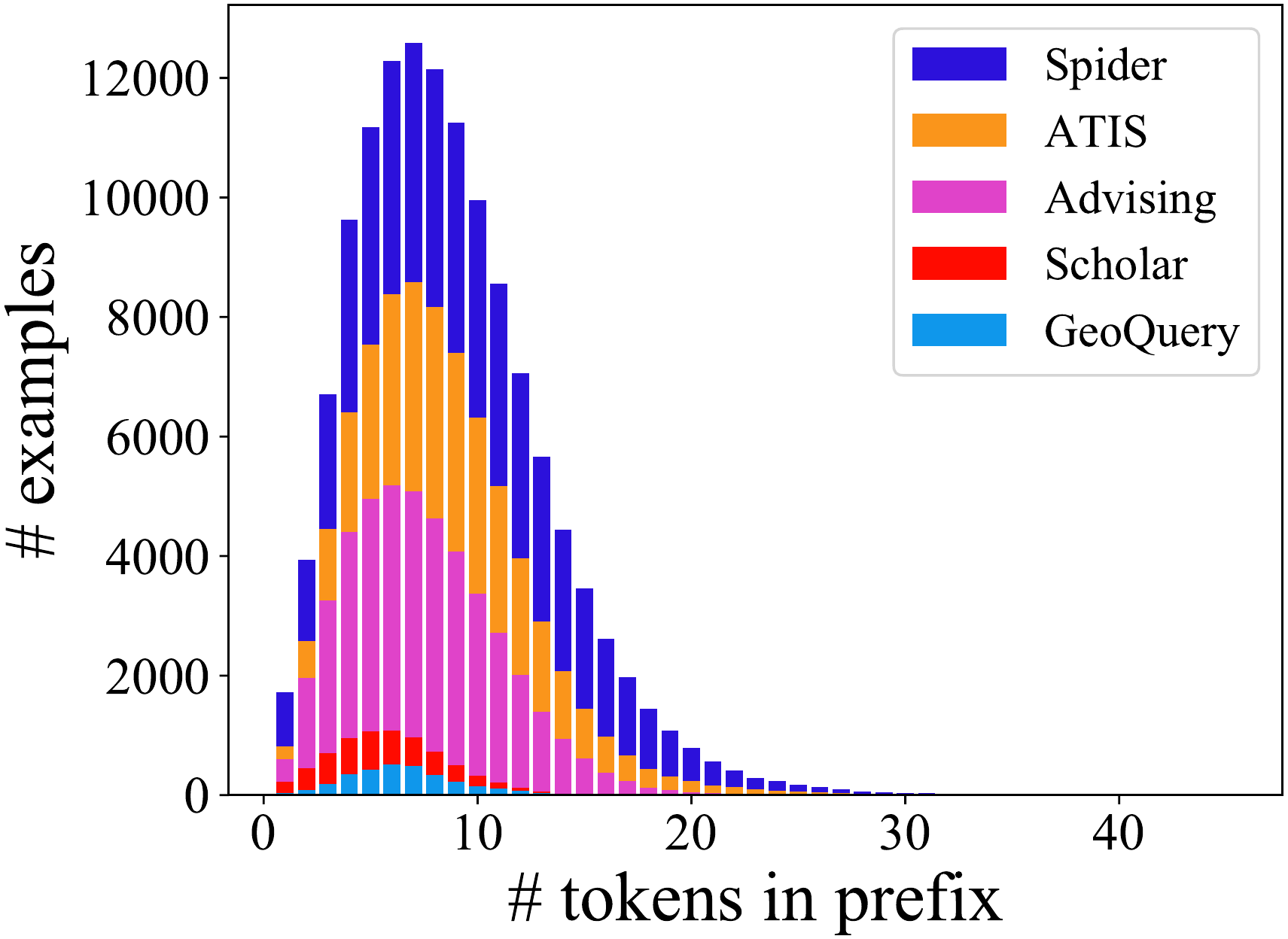}\caption{}
        \label{subfig:query_prefix_len_distribution}
    \end{subfigure}
    
    \begin{subfigure}[c]{0.4\textwidth}
        \includegraphics[width=\textwidth]{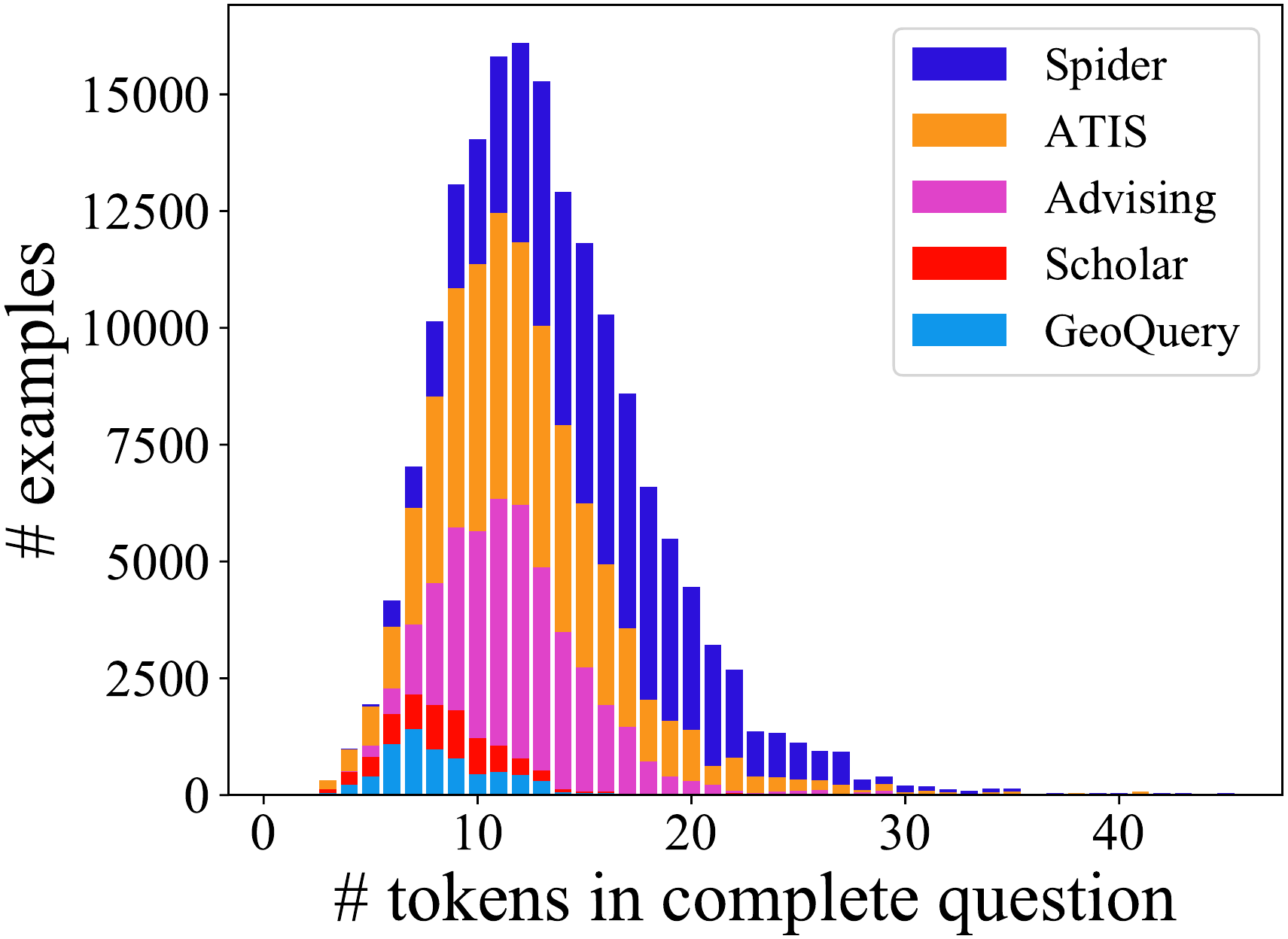}\caption{}
        \label{subfig:question_original_len_distribution}
    \end{subfigure}\hspace{5mm}
    ~ 
    \begin{subfigure}[c]{0.4\textwidth}
    \includegraphics[width=\textwidth]{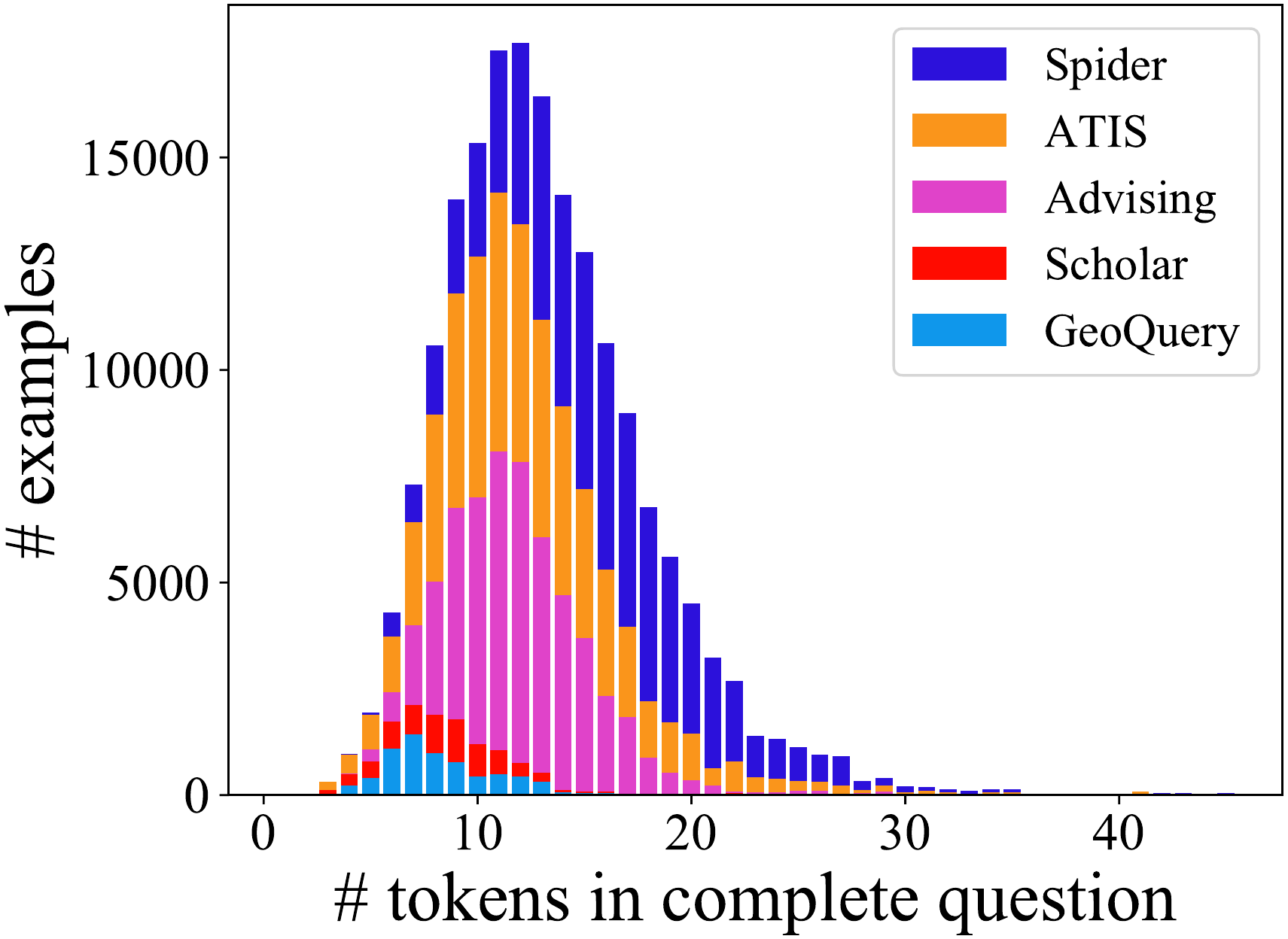}\caption{}
        \label{subfig:query_original_len_distribution}
    \end{subfigure}
    \caption{Prefix length (token count in prefix) distribution for question split (a) and SQL query split (b), and complete questions' length (token count in complete questions) distribution for question split (c) and SQL query split (d).}
    \label{fig: len-distribution}
\end{figure*}

\begin{figure*}[ht!]
    \centering
   \begin{subfigure}[c]{0.3\textwidth}
        \includegraphics[width=\textwidth]{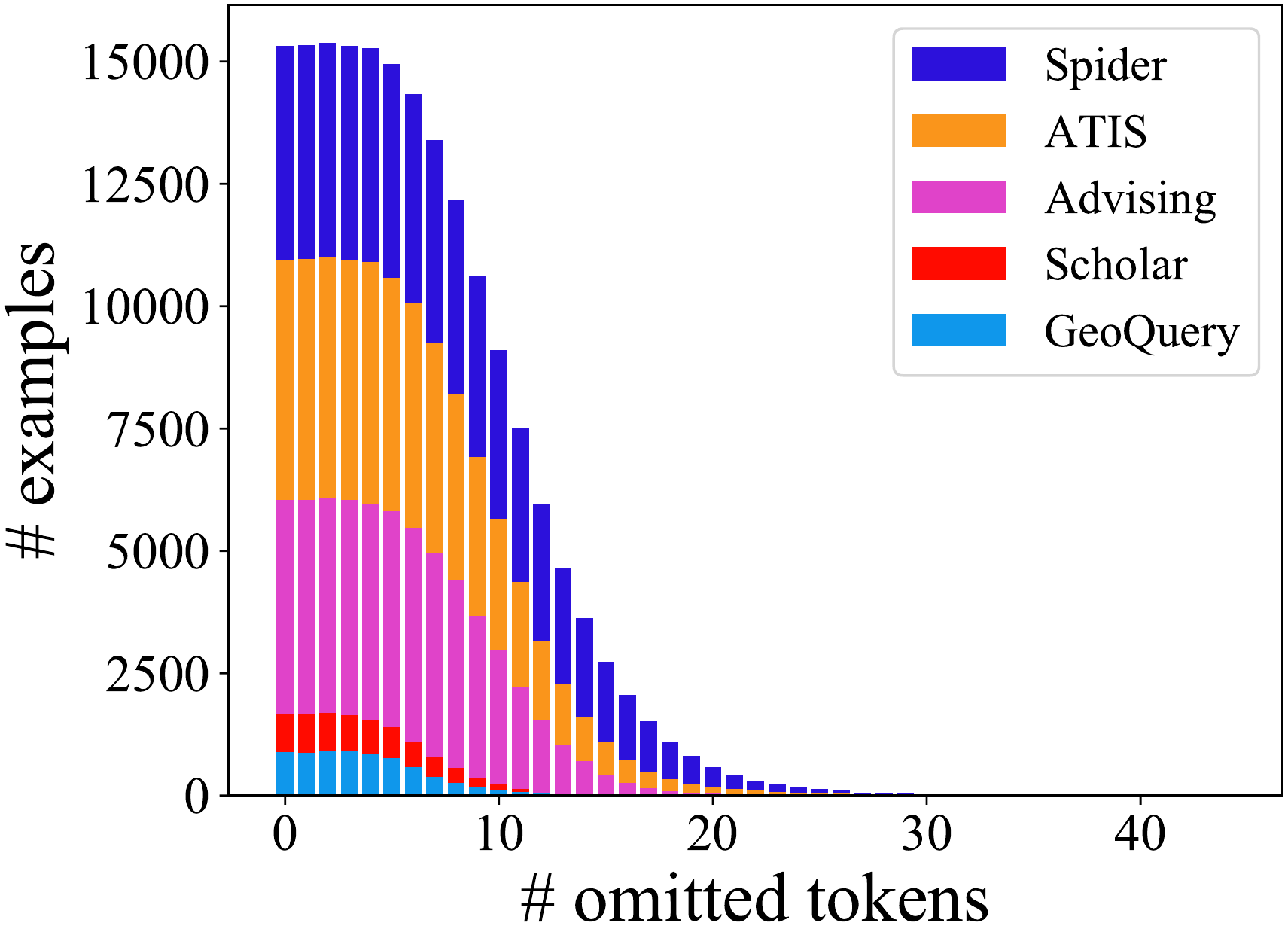}\caption{}
        \label{subfig:query_missing_len_distribution}
    \end{subfigure}\hspace{5mm}
    ~ 
    \begin{subfigure}[c]{0.3\textwidth}
         \includegraphics[width=\textwidth]{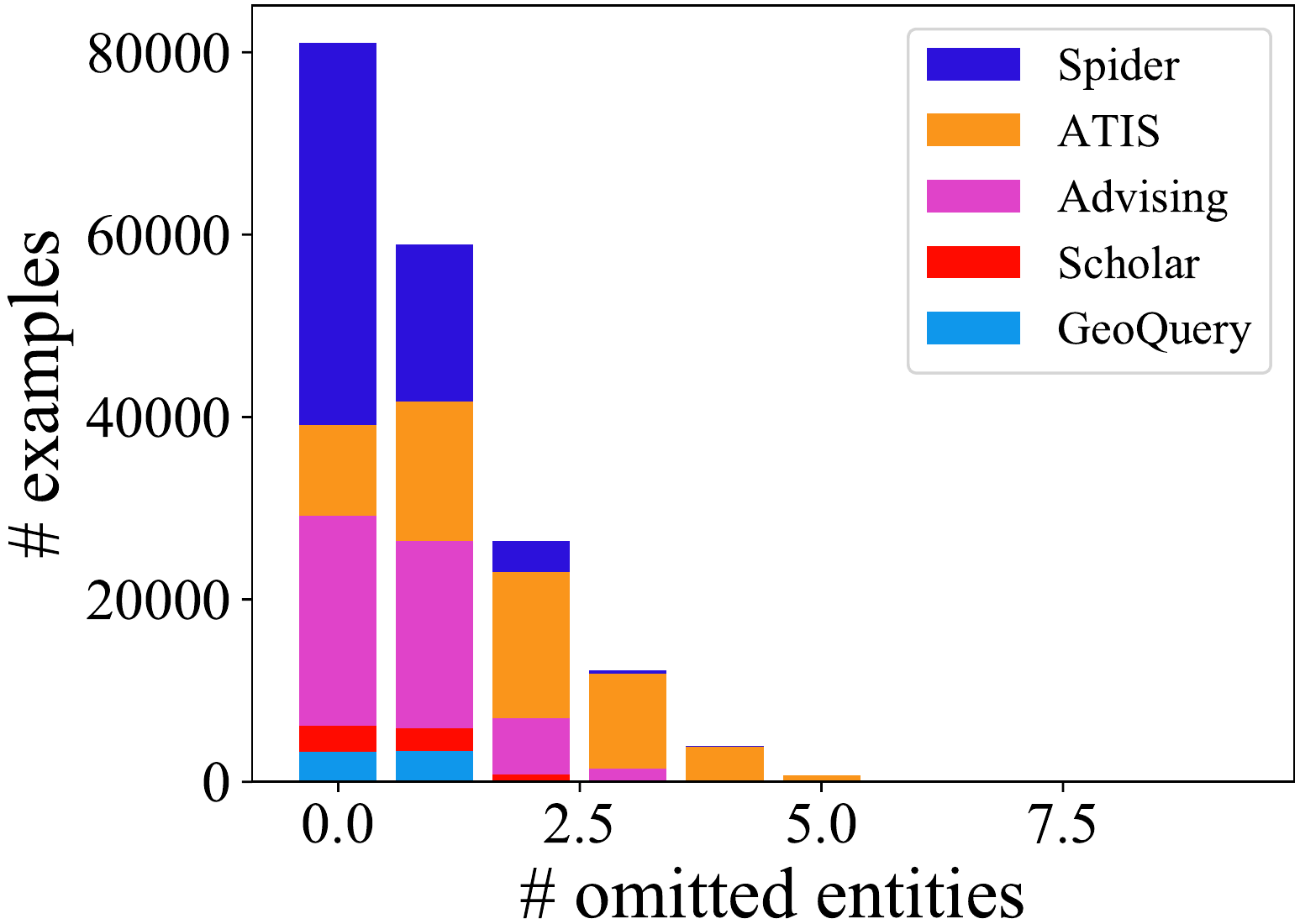}
         \caption{}
        \label{subfig:query_missing_ents}
    \end{subfigure}\hspace{5mm}
    ~
    \begin{subfigure}[c]{0.3\textwidth}
         \includegraphics[width=\textwidth]{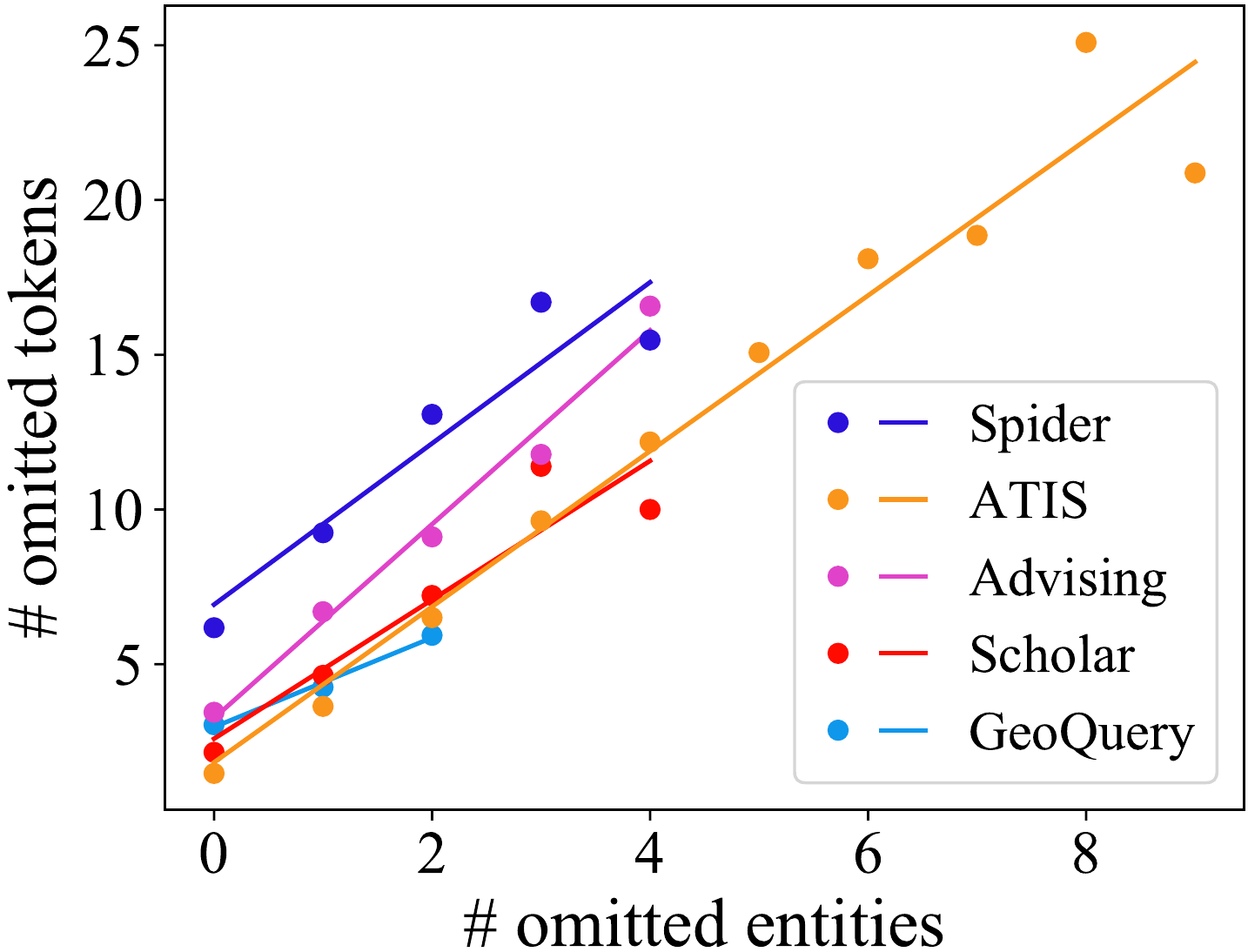}
         \caption{}
        \label{subfig:query_omit_len_ent_ratio}
    \end{subfigure}\hspace{5mm}
    
    \caption{Stacked distribution for (a) the number of omitted tokens from complete question and (b) the number of omitted entities for SQL query split; (c) the number of omitted tokens (vertical axis) v.s. the number of omitted entities (horizontal axis) for SQL query split. The horizontal axis represents (a) the number of omitted tokens and (b) the number of entities in the omitted text for each domain in \datas. The vertical line shows the corresponding number of examples.}
    \label{fig: query-missing-len-info}
\end{figure*}

\begin{figure}[ht!]
    \centering
    \includegraphics[width=0.45\textwidth]{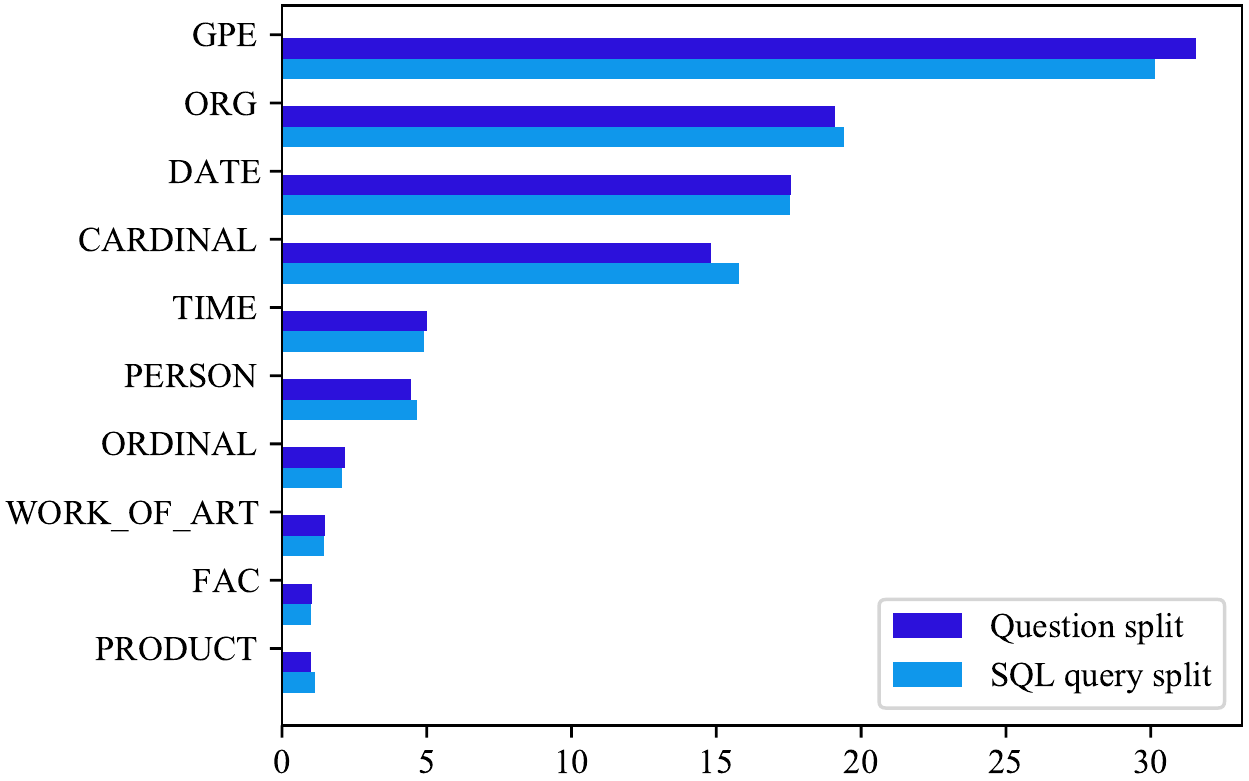}
    \caption{Percentage of types of entities detected by SpaCy~\cite{spacy2} in the omitted text for question split and SQL query split in \datas.}
    \label{fig: ent-percentage}
\end{figure}

\subsection{Dataset Analysis}
\label{appendix: dataset-split-analysis}

% \paragraph{Length Distribution}
Table~\ref{tab:data-stat-8-datasets} report statistics of the generated prefixes and their corresponding SQL queries. Figure~\ref{fig: len-distribution} reports the distribution of prefix length (number of tokens in the prefix) as well as complete question length for the question split and SQL query split. Figure~\ref{subfig:query_missing_len_distribution} and Figure~\ref{subfig:query_missing_ents} show stacked distribution of the number of omitted tokens and the number of omitted entities detected by SpaCy \cite{spacy2} on SQL query split for each sub-task of \datas. Figure~\ref{subfig:query_omit_len_ent_ratio} demonstrates that there is also a linear relationship between the number of omitted tokens and the number of omitted entities for SQL query split, and more entities will be missing if there are more omitted tokens.

\begin{table*}[t]
\small
 \setlength{\tabcolsep}{7pt}
 \centering
     \begin{tabular}{l r r r r r r r r r r r}
     \toprule
     & \multirow{2}{*}{split}  & \multicolumn{2}{c}{Advising} & \multicolumn{2}{c}{ATIS} & \multicolumn{2}{c}{GeoQuery}  & \multicolumn{2}{c}{Scholar} & \multicolumn{2}{c}{Spider}\\
           & & Q &  S & Q & S & Q & S & Q & S & Q & S \\
     \midrule
     \multirow{3}{*}{\# prefixes (P)}  & train  & 23072 & 17812 & 27533 & 29407 & 1784 & 1887 & 2691 & 2119 & 31772 & 31713\\ 
     & dev  & 2279 & 4606 & 3989 & 997 & 253 & 519 & 611 & 667 & 6279 & 6232\\  
     & test & 5369 & 16332 & 2720 & 2953 & 1063 & 633 &1345 & 1625 & 13133 & 13199 \\
    \midrule
     \multirow{3}{*}{\# SQL queries (S)}  & train & 30481 & 23821 & 47375 & 51072 & 4205 & 4304 & 3808 & 2939 & 40208 & 40234 \\  
     & dev  & 2661 & 5801 & 5453 & 1363 & 390 & 1149 & 749 & 793 & 7254 & 7013\\ 
     & test & 6607 & 21558 & 4067 & 3983 & 2210 & 1384 & 1686 & 2308 & 15530 & 15701\\
    \midrule
     \multirow{3}{*}{S/P}
       & train  & 1.32 & 1.34 & 1.72 & 1.74 & 2.36 & 2.28 &1.42 &1.39 & 1.27 & 1.27 \\  & dev  & 1.17 & 1.26 & 1.37 & 1.37 & 1.54 & 2.21 & 1.23 & 1.19 & 1.12 & 1.13 \\  & test & 1.23 & 1.32 & 1.50 & 1.35 & 2.08 & 2.19 & 1.25 & 1.42 &1.18 & 1.19\\
     \bottomrule

     \end{tabular}
 \caption{Statistics for \datas on question splits (``Q'' columns) and SQL query splits (``S'' columns) on train, validation (dev) and test split}
 \label{tab:data-stat-8-datasets}
 \end{table*}

\begin{figure*}[ht!]
    \centering
    \includegraphics[width=\textwidth]{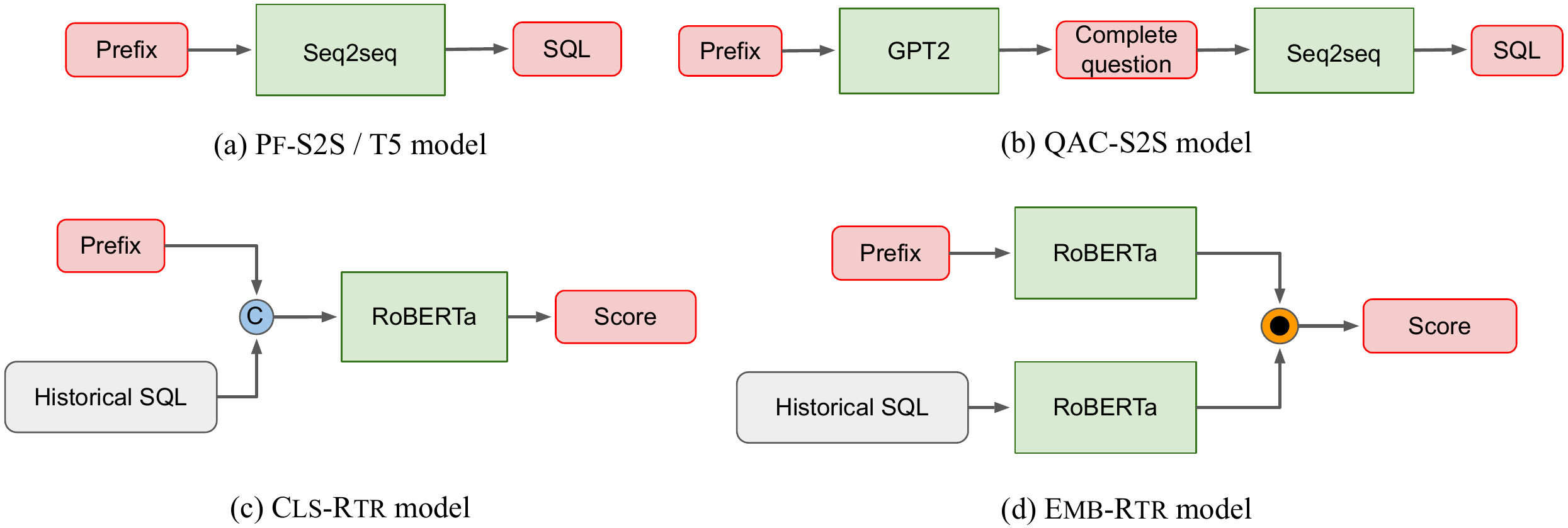}
    \caption{The framework of our generation-based and retrieval-based models. Generation models predict SQL queries based on the input sequence directly, while retrieval models predict a score for each historical SQL and then rank the scores to recommend SQL queries.}
    \label{fig: 4-models}
\end{figure*}

\subsection{Model and Experiment Details}
\label{appendix: model-details}

Figure~\ref{fig: 4-models} shows structures of our baseline models. 

\subsubsection{Model inputs}

For variants of \textsc{Pf-S2S} models, \textsc{T5}, \textsc{Emb-Rtr} and \textsc{Cls-Rtr}, the models' input is question prefix concatenated with the database schema for the prefix.

In the first stage of \textbf{\textsc{QAC-S2S}}, we feed original questions into the model directly to train the GPT-2 language model. For the second stage, we feed the original text-to-SQL dataset to the model.

\subsubsection{\textsc{Pf-S2S} models}
\label{appendix: seq2seq-models}

We implement variations of seq2seq (\textsc{S2S}) models by OpenNMT, a package proposed by \citet{klein-etal-2017-opennmt}. We use Glove embedding \cite{pennington2014glove} for all variations of \textsc{S2S} models we experiment. We use bidirectional LSTM as our encoder, and use LSTM \cite{hochreiter1997long} as our decoder. We have two layers with hidden size as $384$ for decoder. We set learning rate as $10^{-3}$ in our experiments.

\subsubsection{\textsc{QAC-S2S} model}
We finetune distilled version of GPT2 to complete questions based on the prefix. We set the learning rate to $10^{-5}$ and a warm-up ratio of the model to $0.2$. The maximum sequence length is set to $256$. Sequences that are longer than the maximum sequence length will be truncated. We train the \textsc{S2S + Attn + Copy} model on the original text-to-SQL dataset for each of the sub-tasks in \datas. The hyperparameters for the second stage are the same as in Appendix~\ref{appendix: seq2seq-models}. 

During inference, GPT2 first completes the prefix. The completed questions are then fed to \textsc{S2S + Attn + Copy} model. We multiply the probabilities returned by the two stages and select the top $K$ SQL predictions.

\subsubsection{\textsc{T5} model}

We finetune the T5-base\cite{raffel2020exploring} model on each sub-task in \datas. We set the learning rate to $3\cdot10^{-4}$ and the maximum sequence length to $512$.

\subsubsection{\textsc{Emb-Rtr} model}
We finetune RoBERTa-base~\cite{liu2019roberta} encoder model on \datas. Two independent RoBERTa encoders are used to obtain the embeddings of the question prefix and SQL separately. The dot-product of question prefix and SQL embeddings is treated as their similarity score. We treat SQL queries that match a question prefix as the positive examples and sample SQL queries that do not match the prefix as the negative examples in the training process. We make the number of negative examples $5$ times the number of positive examples. We adopt a learning rate of $2\cdot10^{-5}$ to finetune the two RoBERTa encoders. After the training process, all historical SQL embeddings are cached, so we only need to run the RoBERTa encoder once for each given question prefix. We rank all historical SQL queries by the doc-product of their embeddings with the given question prefix embedding.

\subsubsection{\textsc{Cls-Rtr} model}
\label{appendix:roberta-based-sql-retrieval-model}
We use pre-trained RoBERTa-base~\cite{liu2019roberta} as our classification model. The model takes the concatenation of a question prefix and a SQL candidate as input to predict whether the SQL matches the question prefix intention. Given a question prefix during inference, the classification model needs to be run on all historical SQL queries to select the top $K$ SQL queries, which makes the running time significantly larger than all the other models.  Similar to \textbf{\textsc{Emb-Rtr}}, we choose the learning rate as $2\cdot10^{-5}$ and the ratio of negative examples over positive examples is set to 5.

\begin{table*}[t]
\small
\centering
 \begin{tabular}{l   r  r  r  r  r  r  r  r  r  r  r  r  r  r  r  }
     \toprule
     \multirow{2}{*}{Model}  & \multicolumn{3}{c}{Advising} & \multicolumn{3}{c}{ATIS} & \multicolumn{3}{c}{GeoQuery} & \multicolumn{3}{c}{Scholar} & \multicolumn{3}{c}{Spider} \\
    & \multicolumn{1}{c}{R} & \multicolumn{1}{c}{M} & \multicolumn{1}{c}{S} 
     & \multicolumn{1}{c}{R} & \multicolumn{1}{c}{M} & \multicolumn{1}{c}{S}
      & \multicolumn{1}{c}{R} & \multicolumn{1}{c}{M} & \multicolumn{1}{c}{S}
       & \multicolumn{1}{c}{R} & \multicolumn{1}{c}{M} & \multicolumn{1}{c}{S}
        & \multicolumn{1}{c}{R} & \multicolumn{1}{c}{M} & \multicolumn{1}{c}{S} \\
    \midrule
    \textsc{Pf-S2S}   &18 & 14  & 9        &13  & 8   & \textbf{8}    &29  & 21 & 13     & 15 & 11  & \textbf{13}   & \textbf{41} &  27  & 43\\
     
    \textsc{ + Attn}  &19 & 14   & 9       &12 & 8 & 7      &31 &  22  &  15    & \textbf{16} & \textbf{12} &\textbf{12} & 40 & \textbf{29} & 42 \\
    
    \textsc{ + Copy} & \textbf{25}  &  \textbf{19} & \textbf{15}    &11 & 9 &  \textbf{8}      & 28 &  18 &  14     & 12 & 7 &9    & 38 & 24 & \textbf{44}\\
    
    \textsc{QAC-S2S}   & 11 & 7 &  8      & \textbf{14}  & \textbf{11} &  6     & \textbf{40} & \textbf{32} & \textbf{16}   & 12 &  9 & 10 &   -&- & 1\\
    \multicolumn{16}{c}{}\\[-0.8em]
    \hdashline
    \multicolumn{16}{c}{}\\[-0.8em]
    \textsc{T5} &\textbf{49} & \textbf{39} & \textbf{32}     &12 & 8 & 6    & \textbf{40} & \textbf{34} & 13   & \textbf{39} & \textbf{30} &\textbf{28} & 47 & 37 & 43\\
      \midrule
     \textsc{Emb-Rtr} &- & - & -      & -  & - & -            &26 & 19 & 10                   &3 & 1 & 2        & 20 & 6 & 19 \\
          
      \textsc{Cls-Rtr}  & \textbf{23} & \textbf{15} & \textbf{11}                   & \textbf{8} & \textbf{4} & \textbf{5}              & \textbf{31} & \textbf{22} & \textbf{13}             & \textbf{23} & \textbf{15} & \textbf{15}      & \textbf{63} & \textbf{52} & \textbf{61}\\
      \bottomrule
      
     \end{tabular}
\caption[Caption for LOF]{ \textsc{Recall}@10 (R), \textsc{MRR}@10 (M) and \textsc{Save}@10 (S) in percentage for each subtask in \datas on question split. We embolden the best scores for baselines other than T5 above and below the dashed line, respectively. For T5, we embolden its score if it is the highest among all the baselines. We use ``-'' to denote scores $<1\%$.}
 \label{tab: recall-10-generative-models}
 \end{table*}

\begin{table}[t]
 \small
 \setlength{\tabcolsep}{3pt}
 \centering

    \begin{tabular}{l   r  r  r r  r  r  r  r  r  r  r  r }
     \toprule
     \multirow{2}{*}{Model}  & \multicolumn{3}{c}{Advising} & \multicolumn{3}{c}{ATIS} & \multicolumn{3}{c}{GeoQuery} & \multicolumn{3}{c}{Spider} \\
      & \multicolumn{1}{c}{R} & \multicolumn{1}{c}{M} & \multicolumn{1}{c}{S} 
    & \multicolumn{1}{c}{R} & \multicolumn{1}{c}{M} & \multicolumn{1}{c}{S} 
     & \multicolumn{1}{c}{R} & \multicolumn{1}{c}{M} & \multicolumn{1}{c}{S}
      & \multicolumn{1}{c}{R} & \multicolumn{1}{c}{M} & \multicolumn{1}{c}{S}\\
        \midrule
    
    \textsc{Pf-S2S}  & - & - & -             & \textbf{4} & 1 &  \textbf{2}               & 10 &4 &  4                 & 3 & 1 & 4\\
     
    \textsc{ + Attn}  & - & - & -            & \textbf{4} & 1 & \textbf{2}            & \textbf{11} & \textbf{5} & 4         & \textbf{4} & 1 & 4\\
    
    \textsc{ + Copy}  & - & - & -            & 2 & 1 &  1                     & 10 &3 & 4                  & 3 & 1 & 4 \\
    
    \textsc{QAC-S2S}  & - &- & -             & \textbf{4} & \textbf{2} &  \textbf{2}                      & \textbf{11} & \textbf{5} &  4                 &  - & - & -\\
    \multicolumn{6}{c}{}\\[-0.8em]
    \hdashline
    \multicolumn{6}{c}{}\\[-0.8em]
    \textsc{T5} & \textbf{2} & - & \textbf{1}            & 2 & 1 & 1                              & \textbf{15} & \textbf{10} & \textbf{5}           & \textbf{10} & \textbf{6} & \textbf{9}\\

    \bottomrule
     
     \end{tabular}

 \caption[Caption for LOF]{\textsc{Recall}@10 (R), \textsc{MRR}@10 (M) and \textsc{Save}@10 (S) in percentage generative models on SQL query split. We omit the results for sub-tasks Scholar because models perform $0-1\%$ for all metrics. We use ``-'' to denote scores $<1\%$.}
 \label{tab: scores-query-split-generative-10}
 \end{table}

 \begin{figure*}[ht!]
    \centering
   \begin{subfigure}[c]{0.3\textwidth}
        \includegraphics[width=\textwidth]{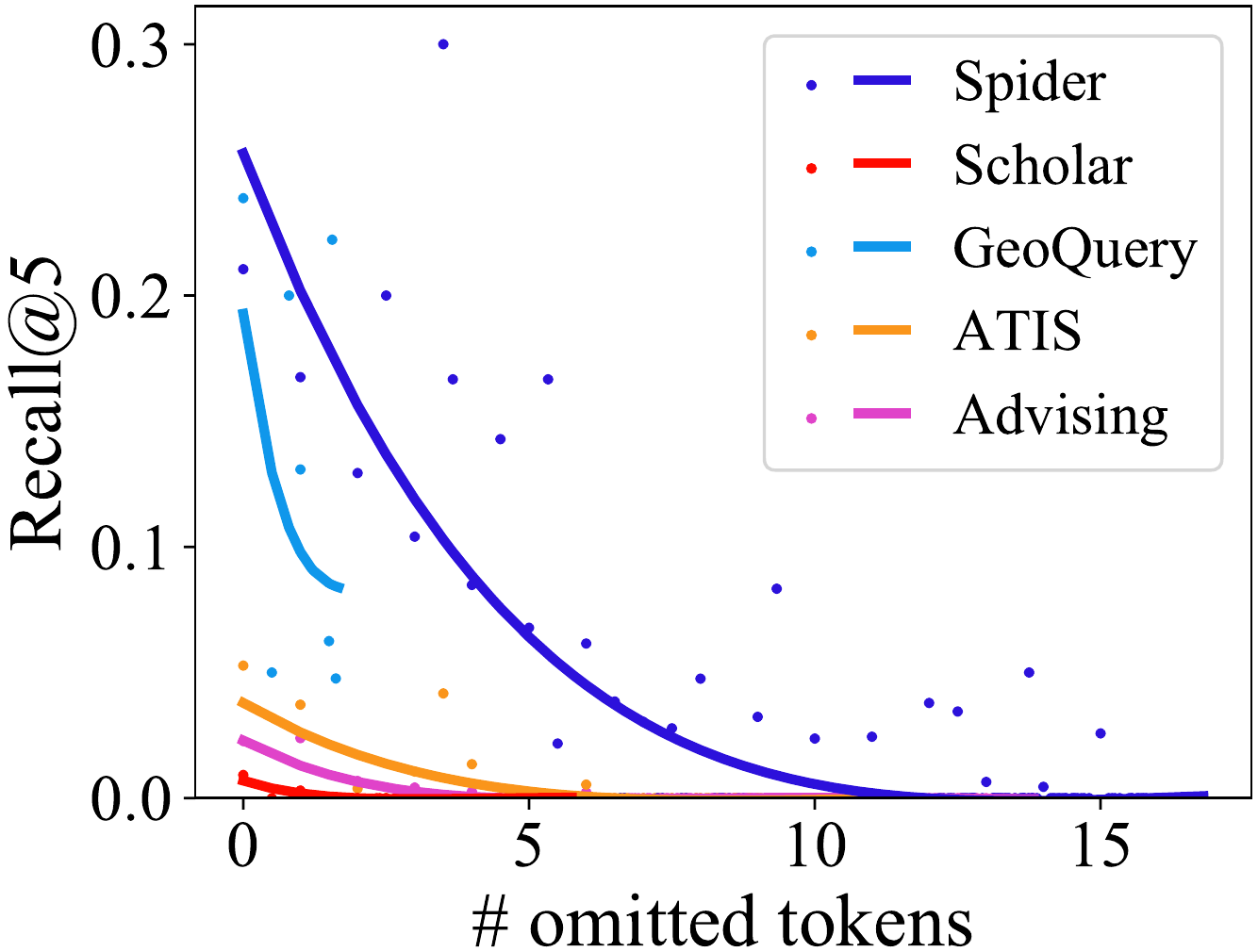}\caption{}
    \end{subfigure}\hspace{5mm}
    ~ 
    \begin{subfigure}[c]{0.3\textwidth}
    \includegraphics[width=\textwidth]{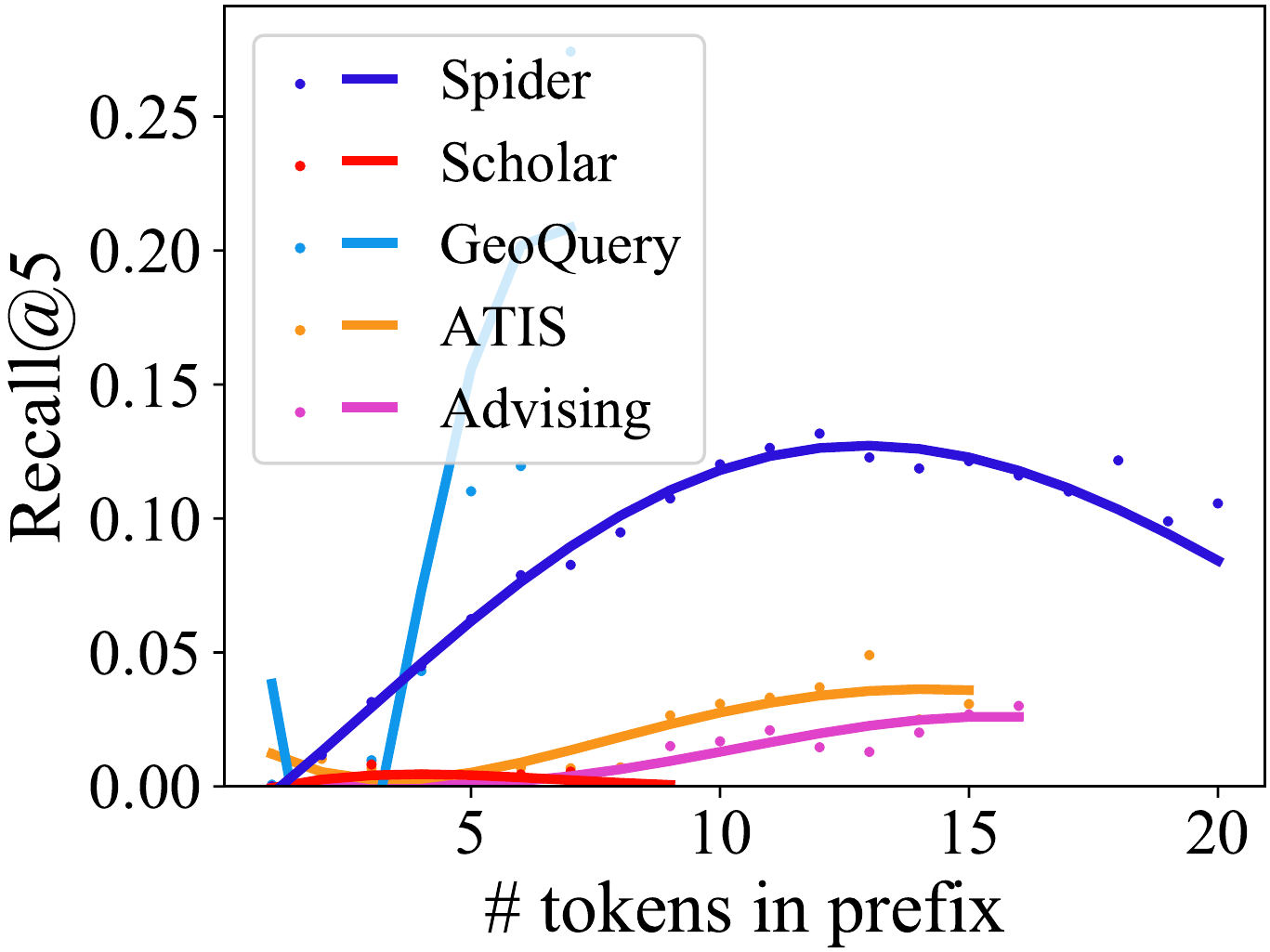}\caption{}
    \end{subfigure}
     ~ 
    \begin{subfigure}[c]{0.3\textwidth}
    \includegraphics[width=\textwidth]{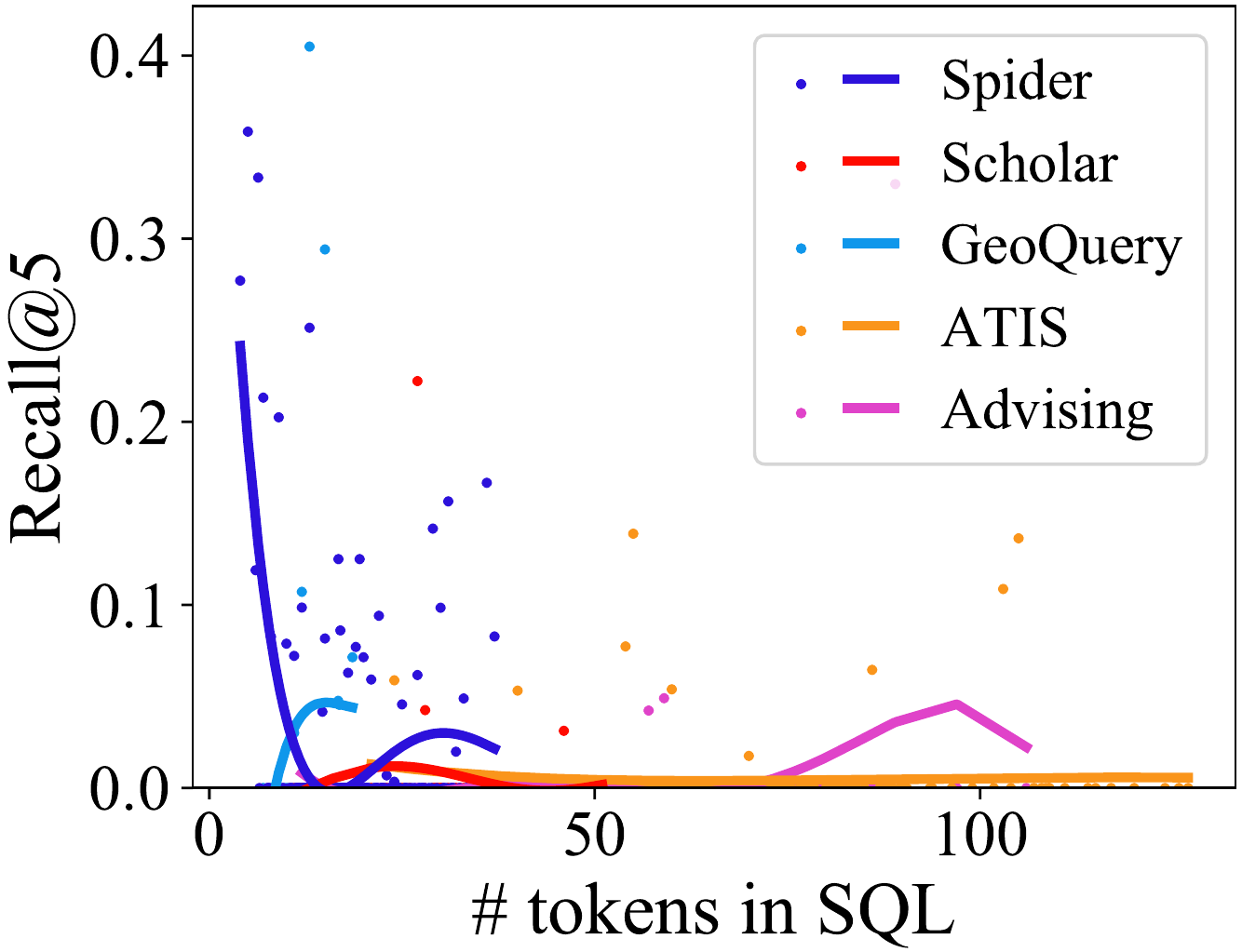}\caption{}
    \end{subfigure}
    \caption{T5's (a) \textsc{Recall}@5 v.s. number of omitted tokens, (b) \textsc{Recall}@5 v.s. prefix length (number of tokens in prefix), (c) \textsc{Recall}@5 v.s. SQL length (number of tokens in SQL) on SQL query split for each sub-task in \datas. We plot \textsc{Recall}@5 corresponding to $\ge 50$ and $\ge 100$ examples for sub-tasks other than Spider and Spider, respectively. \textsc{Recall}@5 is negatively correlated with number of omitted tokens and possesses no monotonic relationships with either prefix length or SQL length.}
    \label{fig: perform-different-setting-query-split}
\end{figure*}

\subsection{Supplementary Results}
\label{appendix: other-metrics}

Table~\ref{tab: recall-10-generative-models} reports models' \textsc{Recall}@10, \textsc{MRR}@10 and \textsc{Save}@10 scores in percentage for each sub-task in \datas on question split. Table~\ref{tab: scores-query-split-generative-10} reports the scores for generative models on all sub-tasks except Scholar in \datas on SQL query split. Because models only perform $0-1\%$ on all metrics for sub-task Scholar, we do not report the results in Table~\ref{tab: scores-query-split-generative-10}.

Figure~\ref{fig: perform-different-setting-query-split} is similar to Figure~\ref{fig: perform-different-setting} but is on SQL query split. Figure~\ref{fig: perform-different-setting-query-split} also shows that T5's \textsc{Recall}@5 scores are negatively correlated with the length to complete (number of omitted tokens) but possess no monotonic relationships with either the prefix length (number of tokens in the prefix) or SQL length (number of tokens in SQL queries).

\end{document}